\documentclass[10pt]{article} 
\usepackage[preprint]{tmlr}

\usepackage{amsmath,amsfonts,bm}

\def\eqref#1{equation~\ref{#1}}

\def\1{\bm{1}}

\DeclareMathAlphabet{\mathsfit}{\encodingdefault}{\sfdefault}{m}{sl}
\SetMathAlphabet{\mathsfit}{bold}{\encodingdefault}{\sfdefault}{bx}{n}

\usepackage{hyperref}
\usepackage{url}
\usepackage{algorithm}
\usepackage{algorithmic} 
\usepackage{subcaption}
\usepackage{graphicx}
\usepackage{placeins}

\newcommand{\remarkName}{Remark}
\newtheorem{remark}{\remarkName}[section]

\title{Variational autoencoder with weighted samples for high-dimensional non-parametric adaptive importance sampling}

\author{\name Julien Demange-Chryst \email julien.demange-chryst@onera.fr \\
      \addr ONERA/DTIS, Université de Toulouse, F-31055 Toulouse, France\\
      Institut de Mathématiques de Toulouse, UMR5219 CNRS, 31062 Toulouse, France
      \ANDauthor
      \name François Bachoc \email francois.bachoc@math.univ-toulouse.fr \\
      \addr Institut de Mathématiques de Toulouse, UMR5219 CNRS, 31062 Toulouse, France
      \ANDauthor
      \name Jérôme Morio \email jerome.morio@onera.fr\\
      \addr ONERA/DTIS, Université de Toulouse, F-31055 Toulouse, France
      \ANDauthor
      \name Timothé Krauth \email timothe.krauth@zhaw.ch \\
      \addr ONERA/DTIS, Université de Toulouse, F-31055 Toulouse, France\\
      Zurich University of Applied Sciences, Centre for Aviation, Winterthur, Switzerland}

\def\month{MM}  
\def\year{YYYY} 
\def\openreview{\url{https://openreview.net/forum?id=XXXX}} 

\begin{document}

\maketitle

\begin{abstract}
Probability density function estimation with weighted samples is the main foundation of all adaptive importance sampling algorithms. Classically, a target distribution is approximated either by a non-parametric model or within a parametric family. However, these models suffer from the curse of dimensionality or from their lack of flexibility. In this contribution, we suggest to use as the approximating model a distribution parameterised by a variational autoencoder. We extend the existing framework to the case of weighted samples by introducing a new objective function. The flexibility of the obtained family of distributions makes it as expressive as a non-parametric model, and despite the very high number of parameters to estimate, this family is much more efficient in high dimension than the classical Gaussian or Gaussian mixture families. Moreover, in order to add flexibility to the model and to be able to learn multimodal distributions, we consider a learnable prior distribution for the variational autoencoder latent variables. We also introduce a new pre-training procedure for the variational autoencoder to find good starting weights of the neural networks to prevent as much as possible the posterior collapse phenomenon to happen. At last, we explicit how the resulting distribution can be combined with importance sampling, and we exploit the proposed procedure in existing adaptive importance sampling algorithms to draw points from a target distribution and to estimate a rare event probability in high dimension on two multimodal problems.
\end{abstract}

\section{Introduction}

Importance sampling is a well-known uncertainty quantification method which requires to deal with weighted samples, i.e. a set of observations paired with a set of corresponding weights. It is classically used for estimating an expectation, such as a failure probability, in the aim of reducing the variance of the Monte Carlo estimator \cite{Kahn1951SplittingParticleTransmission,rubinstein2004cross,kurtz2013cross}, or for generating points from a target probability distribution \cite{cappe2004population,cornuet2012adaptive,marin2019mamis}. What these algorithms have in common is that they all require to estimate a target probability distribution with weighted samples, and obviously, the accuracy of the algorithm depends on the quality of the estimation of the distribution \cite{chatterjee2018sample}. A first way for this estimation is to use non-parametric models, such as kernel smoothing \cite{wand1994kernel,scott2005multidimensional}. These models are flexible but they strongly suffer from the curse of dimensionality. Another solution is to use parametric families of distributions, such as the Gaussian \cite{rubinstein2004cross} or Gaussian mixture \cite{kurtz2013cross,geyer2019cross} ones, which are more robust in medium-high dimension. However, they sometimes require some prior knowledge on the target distribution, and their lack of flexibility and the huge number of parameters to estimate \cite{au2003important} can negatively impact the quality of the estimation when the dimension is high.

In order to combine both flexibility and robustness to the dimension, we suggest here to use as the approximating model a distribution parameterised by a variational autoencoder, whose main principle has been introduced in the last decade \cite{kingma2014autoencoding,kingma2019introduction}. Variational autoencoders are deep generative models for approximating high-dimensional complex distributions of observed data and generating new samples. The specific feature of a variational autoencoder compared to other density estimation methods is that it performs a dimensionality reduction into a lower dimensional latent space in order to facilitate the estimation. Moreover, in opposition to other dimensionality reduction techniques such as principal component analysis \cite{wold1987principal} or autoencoders \cite{mcclelland1987parallel,bank2023autoencoders}, variational autoencoders have good generation properties and give explicitly the approximating distribution, allowing to perform Monte Carlo simulations. This tool is now popular in the machine leaning community but not so much in uncertainty quantification. In the present article, we extend this framework to probability density estimation with weighted observations, and we apply it to adaptive importance sampling.

The remainder of this paper is organized as follows. First, Section \ref{sec:vae_review} formally presents the problem of density estimation and provides a review on the principle of variational autoencoders and some improvements. Then, Section \ref{sec:new_vae_procedure} introduces our suggested extension of variational autoencoders to weighted samples. In addition, Section \ref{sec:posterior_collapse} presents the posterior collapse phenomenon and a procedure to handle it. Section \ref{sec:numerical_results} illustrates the practical interest of the proposed procedure on the generation from some target distributions and on the estimation of failure probabilities on two multimodal problems. Finally, Section \ref{sec:conclusion} concludes the present article and gives future research perspectives stemming from it.

\section{Probability density estimation supported by variational autoencoders}
\label{sec:vae_review}

Probability density function (PDF) estimation is a major topic of interest in statistics, as the PDF fully characterises the distribution of a continuous random vector. Given an observed dataset $\left(\mathbf{X}^{(n)}\right)_{n\in[\![1,N]\!]}$ of points from the \textit{input space} $\mathcal{X}\subseteq\mathbb{R}^d$ of dimension $d\geq 1$, it consists in the construction of an estimate of the true underlying unknown PDF $g^* : \mathcal{X} \rightarrow \mathbb{R}_+$. There exist several strategies to do so. First, parametric methods consist in picking the best representative of the target density $g^*$ within a parametric family of densities. A few classical families \cite{kotz2004continuous} are often used and can accurately model a fair number of phenomena, but their lack of flexibility and/or the high amount of parameters to estimate when the dimension $d$ increases is a limitation. Second, an important class of non-parametric methods is provided by kernel smoothing \cite{wand1994kernel,scott2005multidimensional}. Despite the larger flexibility of the corresponding approximating models and some improvements \cite{silverman1986density,zhang2006bayesian,perrin2018data}, kernel smoothing strongly suffers from the curse of dimensionality. One can also mention functional decomposition methods \cite{wasserman2006all} or non-parametric Bayesian methods \cite{ghosal2017fundamentals}, but they also suffer from the curse of dimensionality.

A recent strategy to perform probability density estimation in high dimension is offered by variational autoencoders (VAEs). The principle of this probabilistic tool was first introduced in \cite{kingma2014autoencoding}. We refer to \cite{kingma2019introduction} for an exhaustive and up-to-date presentation. Let us then present in this section the general principle of VAEs.

\subsection{Latent variable and Bayesian variational inference}
\label{ssec:latent_var}

Bayesian variational inference \cite{fox2012tutorial} consists in approximating the target distribution $g^*$ by a parametric model $g_{\boldsymbol{\theta}}$, where $\boldsymbol{\theta}$ denotes its parameters. In order to facilitate the density estimation, Bayesian variational inference introduces an unobserved \textit{latent variable} $\mathbf{z}$ which lies in a lower-dimensional space $\mathcal{Z}\subseteq\mathbb{R}^{d_z}$, with $d_z\ll d$. Together, both variables define a joint distribution on $\mathcal{X}\times\mathcal{Z}$. By marginalising over the latent variable $\mathbf{z}$, the distribution on $\mathcal{X}$ is given by: \begin{equation}\label{eq:g_theta_integral}
    g_{\boldsymbol{\theta}}\left(\mathbf{x}\right) = \int_{\mathcal{Z}}g_{\boldsymbol{\theta}}\left(\mathbf{x},\mathbf{z}\right)d\mathbf{z}.
\end{equation} Nevertheless, the marginalisation over an unknown and unobserved latent variable makes the computation of the integral intractable, and as a consequence also the direct computation of the estimated distribution $g_{\boldsymbol{\theta}}$. However, thanks to Bayes theorem, it is possible to write: \begin{equation}\label{eq:bayes}
    g_{\boldsymbol{\theta}}\left(\mathbf{z}\left|\mathbf{x}\right.\right) = \dfrac{p\left(\mathbf{z}\right)g_{\boldsymbol{\theta}}\left(\mathbf{x}\left|\mathbf{z}\right.\right)}{g_{\boldsymbol{\theta}}\left(\mathbf{x}\right)},
\end{equation} where $p$ is the prior distribution on $\mathbf{z}$ and where $g_{\boldsymbol{\theta}}\left(\mathbf{x}\left|\mathbf{z}\right.\right)$ is the likelihood. Being able to infer the \textit{true posterior} $g_{\boldsymbol{\theta}}\left(\mathbf{z}\left|\mathbf{x}\right.\right)$ allows then to compute $g_{\boldsymbol{\theta}}\left(\mathbf{x}\right)$ from \eqref{eq:bayes}. To do so, we approximate the true posterior by a \textit{variational posterior distribution} $q_{\boldsymbol{\phi}}\left(\mathbf{z}\left|\mathbf{x}\right.\right)$ chosen within an expressive parametric family of distributions parameterized by $\boldsymbol{\phi}$. 

The choice of the dimension of the latent space $d_z$ has a huge influence on the accuracy of the density approximation and should be the best trade-off between dimensionality reduction and loss of information. Indeed, performing the variational inference in a lower dimensional subspace $\mathcal{Z}$ instead of in the high-dimensional input space $\mathcal{X}$ reduces the number of parameters of $q_{\boldsymbol{\phi}}\left(\mathbf{z}\left|\mathbf{x}\right.\right)$ to estimate and makes the process more accurate. However, the dimension of the latent space must be large enough to correctly catch the structure of the data, and thus the structure of the true underlying distribution $g^*$.

\subsection{General principle of VAEs}

In order to perform the density estimation, a VAE consists in using two neural networks, a probabilistic encoder denoted $E_{\boldsymbol{\phi}}$ parameterized by the weights $\boldsymbol{\phi}$, and a probabilistic decoder denoted $D_{\boldsymbol{\theta}}$ parameterized by the weights $\boldsymbol{\theta}$, to model respectively the variational posterior distribution and the likelihood function. Hence, in the case of VAEs, the parameters $\boldsymbol{\phi}$ and $\boldsymbol{\theta}$ discussed in Section \ref{ssec:latent_var} are the weights of $E_{\boldsymbol{\phi}}$ and $D_{\boldsymbol{\theta}}$.

First, the probabilistic encoder performs the dimensionality reduction described above. Indeed, given an input point from the input space $\mathbf{x}\in\mathcal{X}$, it returns the parameters of the approximating variational parametric distribution $q_{\boldsymbol{\phi}}\left(\mathbf{z}\left|\mathbf{x}\right.\right)$. Classically, a Gaussian distribution with diagonal covariance matrix is considered as the variational posterior distribution, such that $q_{\boldsymbol{\phi}}\left(.\left|\mathbf{x}\right.\right) \sim \mathcal{N}\left(\boldsymbol{\mu}_{\mathbf{x}},\boldsymbol{\Sigma}_{\mathbf{x}}\right)$, where its parameters $\left(\boldsymbol{\mu}_{\mathbf{x}},\boldsymbol{\Sigma}_{\mathbf{x}}\right) = E_{\boldsymbol{\phi}}\left(\mathbf{x}\right)$ are the output of the encoder.

Second, the probabilistic decoder sets the parameters of the likelihood function $g_{\boldsymbol{\theta}}\left(\mathbf{x}\left|\mathbf{z}\right.\right)$. Indeed, it maps a point from the latent space $\mathbf{z}\in\mathcal{Z}$ into the corresponding parameters of the likelihood. Once more, for continuous data, the most classical choice for the likelihood is the Gaussian distribution with diagonal covariance matrix. Thus, the likelihood distribution is given by $g_{\boldsymbol{\theta}}\left(.\left|\mathbf{z}\right.\right) \sim \mathcal{N}\left(\boldsymbol{\mu}_{\mathbf{z}},\boldsymbol{\Sigma}_{\mathbf{z}}\right)$, where its parameters $\left(\boldsymbol{\mu}_{\mathbf{z}},\boldsymbol{\Sigma}_{\mathbf{z}}\right) = D_{\boldsymbol{\theta}}\left(\mathbf{z}\right)$ are the output of the decoder. As a result, according to \eqref{eq:g_theta_integral} and with $g_{\boldsymbol{\theta}}\left(\mathbf{x},\mathbf{z}\right) = p\left(\mathbf{z}\right)g_{\boldsymbol{\theta}}\left(\mathbf{x}\left|\mathbf{z}\right.\right)$, the estimated distribution  $g_{\boldsymbol{\theta}}$ can be seen as an infinite mixture of Gaussian distributions.

Even if the distribution estimated by a VAE belongs theoretically to a parametric family, its flexibility, its ability to approximate complex distributions and its form make it closer to a non-parametric model. As discussed at the beginning of Section \ref{sec:vae_review}, other existing non-parametric models strongly face the curse of dimensionality whereas, as shown in the numerical results in Section \ref{sec:numerical_results}, VAEs are much more efficient in high dimension.

The main challenge now is to train both neural networks. The most common method for this is to minimize the Kullback-Leibler divergence \cite{kullback1951information} $D_{\text{KL}}\left(g^*\Vert g_{\boldsymbol{\theta}}\right)$ between the target distribution $g^*$ and the candidate one $g_{\boldsymbol{\theta}}$ w.r.t. the parameters $\boldsymbol{\theta}$. One can show that it is equivalent to maximise the expectation $\mathbb{E}_{g^*}\left[\log\left(g_{\boldsymbol{\theta}}\left(\mathbf{X}\right)\right)\right]$. This is also called cross-entropy minimisation \cite{rubinstein2004cross}. Nevertheless, the computation of this expectation requires the computation of the integral in \eqref{eq:g_theta_integral} which is intractable. However, for fixed $\mathbf{x}\in\mathcal{X}$, one can find a more convenient and easier to compute lower bound of the log-likelihood using the latent variable $\mathbf{z}$ \cite{kingma2014autoencoding,kingma2019introduction}: \begin{equation}\label{eq:likelihood_lower_bound}
    \log\left(g_{\boldsymbol{\theta}}\left(\mathbf{x}\right)\right) \geq \mathbb{E}_{q_{\boldsymbol{\phi}}\left(.|\mathbf{x}\right)}\left[\log\left(g_{\boldsymbol{\theta}}\left(\mathbf{x}|\mathbf{Z}\right)\right)\right] - D_{\text{KL}}\left(q_{\boldsymbol{\phi}}\left(\mathbf{Z}|\mathbf{x}\right)\Vert p\left(\mathbf{Z}\right)\right).
\end{equation}Taking the expectation of $\mathbf{x}$ w.r.t. $g^*$ leads to the objective function of the VAE given by:\begin{equation}\label{eq:elbo}
\mathbb{E}_{g^*}\left[\log\left(g_{\boldsymbol{\theta}}\left(\mathbf{X}\right)\right)\right] \geq \mathbb{E}_{g^*}\left[\mathbb{E}_{q_{\boldsymbol{\phi}}\left(.|\mathbf{X}\right)}\left[\log\left(g_{\boldsymbol{\theta}}\left(\mathbf{X}|\mathbf{Z}\right)\right)\right]\right] - \mathbb{E}_{g^*}\left[D_{\text{KL}}\left(q_{\boldsymbol{\phi}}\left(\mathbf{Z}|\mathbf{X}\right)\Vert p\left(\mathbf{Z}\right)\right)\right] := \text{ELBO}\left(\boldsymbol{\phi},\boldsymbol{\theta}\right).
\end{equation} The training procedure of the VAE aims then to maximize the $\text{ELBO}$ (\textit{Evidence lower bound}) function according to the weights $\left(\boldsymbol{\phi},\boldsymbol{\theta}\right)$. In practice, all the terms of this objective function are estimated using the observed sample $\left(\mathbf{X}^{(n)}\right)_{n\in[\![1,N]\!]}$ distributed according to $g^*$. More precisely, the $\text{ELBO}$ function combines two opposite phenomena, and optimising it consists in finding the best trade-off between them. \begin{itemize}
    \item First, maximising the reconstruction term $\mathbb{E}_{q_{\boldsymbol{\phi}}\left(.|\mathbf{x}\right)}\left[\log\left(g_{\boldsymbol{\theta}}\left(\mathbf{x}|\mathbf{Z}\right)\right)\right]$ in order to have an accurate reconstruction. More precisely, the encoded distributions $q_{\boldsymbol{\phi}}\left(.|\mathbf{X}^{(n)}\right)$ associated to each point from the dataset must be separated enough from each others in the latent space such that the decoder is able to reconstruct the correct point $\mathbf{X}^{(n)}$ from it.
    \item Second, minimising the regularisation term, which is the Kullback-Leibler divergence $D_{\text{KL}}\left(q_{\boldsymbol{\phi}}\left(\mathbf{Z}|\mathbf{x}\right)\Vert p\left(\mathbf{Z}\right)\right)$ between the prior and the variational posterior distribution over the dataset. Indeed, the decoder is trained from samples distributed as $q_{\boldsymbol{\phi}}\left(.|\mathbf{X}^{(n)}\right)$ for all $n\in[\![1,N]\!]$. Then, when new $\mathbf{x}$ samples are generated, the decoder is applied to new $\mathbf{z}$ samples distributed according to the prior $p$. If this prior is too far away from all the $q_{\boldsymbol{\phi}}\left(.|\mathbf{X}^{(n)}\right)$, the new generated points may not be representative of the true underlying distribution $g^*$, because the decoder is applied outside of its training regime.
\end{itemize} 

\subsection{Choice of the prior and VampPrior}

Let us discuss the choice of the prior distribution $p$ on $\mathbf{z}$. Classically, the most common choice for the prior is the standard Normal distribution in dimension $d_z$. This simple choice is practically convenient, especially because it gives an analytical expression of the Kullback-Leibler divergence $D_{\text{KL}}\left(q_{\boldsymbol{\phi}}\left(\mathbf{Z}|\mathbf{x}\right)\Vert p\left(\mathbf{Z}\right)\right)$ since $q_{\boldsymbol{\phi}}\left(\mathbf{z}|\mathbf{x}\right)$ is also Gaussian. However, it is not optimal in general, for multimodal problems for example, and can lead to over-regularisation and consequently to poor density estimation performances. A more flexible prior is most of the time required.

Theoretically, by rewriting the ELBO function with two regularisation terms and by maximising it w.r.t. the prior $p$ \cite{makhzani2015adversarial,hoffman2016elbo}, the optimal prior is analytically given by: \begin{equation}\label{eq:aggregated_post}
    p^*\left(\mathbf{z}\right) = \int_{\mathcal{X}}q_{\boldsymbol{\phi}}\left(\mathbf{z}|\mathbf{x}\right)g^*\left(\mathbf{x}\right)d\mathbf{x} = \mathbb{E}_{g^*}\left[q_{\boldsymbol{\phi}}\left(\mathbf{z}|\mathbf{X}\right)\right].
\end{equation} This optimal prior distribution is called the \textit{aggregated posterior}. One can note that the aggregated posterior is the continuous mixture of all the variational posterior distributions over the whole input space. In particular, it depends on the parameters of the encoder $\boldsymbol{\phi}$. A natural way to approximate $p^*$ is $p^{\text{emp}}\left(\mathbf{z}\right) = \frac{1}{N}\sum_{n=1}^Nq_{\boldsymbol{\phi}}\left(\mathbf{z}|\mathbf{X}^{(n)}\right)$. Nevertheless, since $N$ is usually large, this empirical distribution is intractable in practice and can also lead to over-fitting. To overcome these issues, there are several types of priors to approximate the optimal one $p^*$ \cite{dilokthanakul2016deep,nalisnick2016approximate,kim2018disentangling,chen2018isolating,lavda2019improving,takahashi2019variational,kalatzis2020variational,negri2022meta}, each of them with advantages and drawbacks. Here, we use the \textit{Variational Mixture of Posteriors} prior, or \textit{VampPrior}, introduced in \cite{tomczak2018vae}. It consists in approximating the optimal prior by a mixture distribution of the form: \begin{equation}\label{eq:vampprior}
    p_{\mathbf{u}_1,\dots,\mathbf{u}_K,\boldsymbol{\phi}}^{\text{VP}}\left(\mathbf{z}\right) = \dfrac{1}{K}\sum_{k=1}^Kq_{\boldsymbol{\phi}}\left(\mathbf{z}|\mathbf{u}_k\right),
\end{equation} where $K\geq 1$ is the number of components of the mixture chosen by the user and where the points $\left(\mathbf{u}_k\right)_{k\in[\![1,K]\!]}\in\mathcal{X}$ are learnable pseudo-inputs from the input space. More precisely, the VampPrior distributions with $K$ components constitute a parametric family of distributions parameterised by $\left(\mathbf{u}_1,\dots,\mathbf{u}_K,\boldsymbol{\phi}\right)$. 

Practically, a single neural network denoted $\text{VP}_{\boldsymbol{\lambda}}$ and parameterised by the weights $\boldsymbol{\lambda}$ maps the $K$ vectors of the canonical basis $\left(e_k^K\right)_{k\in[\![1,K]\!]}$ of $\mathbb{R}^K$ into the pseudo-inputs $\left(\mathbf{u}_k\right)_{k\in[\![1,K]\!]}$. Therefore, the VampPrior distribution is only parameterised by the weights $\boldsymbol{\lambda}$ and $\boldsymbol{\phi}$. This new neural network has to be trained in order to find the best pseudo-inputs that maximise the performances of the VAE, and these new parameters have then to appear in the ELBO function which thus becomes: \begin{equation}\label{eq:elbo_vp}
     \text{ELBO}\left(\boldsymbol{\phi},\boldsymbol{\theta},\boldsymbol{\lambda}\right) = \mathbb{E}_{g^*}\left[\mathbb{E}_{q_{\boldsymbol{\phi}}\left(.|\mathbf{X}\right)}\left[\log\left(g_{\boldsymbol{\theta}}\left(\mathbf{X}|\mathbf{Z}\right)\right)\right]\right] - \mathbb{E}_{g^*}\left[D_{\text{KL}}\left(q_{\boldsymbol{\phi}}\left(\mathbf{Z}|\mathbf{X}\right)\Vert p_{\boldsymbol{\lambda},\boldsymbol{\phi}}\left(\mathbf{Z}\right)\right)\right],
\end{equation} where $p_{\boldsymbol{\lambda},\boldsymbol{\phi}}^{\text{VP}}\left(\mathbf{z}\right) = \dfrac{1}{K}\sum_{k=1}^Kq_{\boldsymbol{\phi}}\left(\mathbf{z}|\text{VP}_{\boldsymbol{\lambda}}\left(\mathbf{e}_k^K\right)\right)$. The VampPrior is flexible enough to be adapted to many kinds of problems and has the advantage to depend on the weights of the encoder $\boldsymbol{\phi}$, as in the form of the optimal prior in \eqref{eq:aggregated_post}.

\section{Variational autoencoder with weighted samples}
\label{sec:new_vae_procedure}

In this section, we present a new procedure to estimate, with a VAE, a target probability density $g^*$, no longer by using observations drawn from $g^*$ itself, but rather by using observations drawn from another probability density $f:\mathcal{X}\rightarrow\mathbb{R}_+$. To the best of our knowledge, this problem has only been investigated in \cite{wang2019adaptive} (see Remark \ref{rk:comparison} for a comparison). To that end, we show how to adapt the VAE framework presented in Section \ref{sec:vae_review} to address this issue.




In the context of the rest of the paper, we assume that we have at our disposal a dataset $\left(\mathbf{X}^{(n)}\right)_{n\in[\![1,N]\!]}\in\mathcal{X}^N$ distributed according to a distribution $f:\mathcal{X}\rightarrow \mathbb{R}_+$ known up to a constant, and that we aim to estimate the target probability density $g^*$ also known up to a constant. More precisely, the distributions $f$ and $g^*$ are given by: \begin{equation}
    \left\{
    \begin{array}{rcl}
        f\left(\mathbf{x}\right) &=& \widetilde{f}\left(\mathbf{x}\right)/c_f \\
        g^*\left(\mathbf{x}\right) &=& \widetilde{g}^*\left(\mathbf{x}\right)/c_g,
    \end{array}
\right.
\end{equation} where $\widetilde{f}$ and $\widetilde{g}^*$ are fully known non-negative functions and where $c_f$ and $c_g$ are positive constants. At last, we also assume that the support of $f$ contains the support of $g^*$.

In the same way as in Section \ref{sec:vae_review}, in order to approximate as accurately as possible the target distribution $g^*$, we aim to minimise the Kullback-Leibler divergence $D_{\text{KL}}\left(g^*\Vert g_{\boldsymbol{\theta}}\right)$ w.r.t. the parameters $\boldsymbol{\theta}$, which is equivalent to maximise $\mathbb{E}_{g^*}\left[\log\left(g_{\boldsymbol{\theta}}\left(\mathbf{X}\right)\right)\right]$. However, we have here a sample distributed according to $f$ and it is not possible to compute this expectation under $g^*$. Fortunately, thanks to the importance sampling trick \cite{Kahn1951SplittingParticleTransmission}, it is possible to rewrite it as an expectation under $f$: \begin{equation}
\mathbb{E}_{g^*}\left[\log\left(g_{\boldsymbol{\theta}}\left(\mathbf{X}\right)\right)\right] = \mathbb{E}_{f}\left[\dfrac{g^*\left(\mathbf{X}\right)}{f\left(\mathbf{X}\right)}\log\left(g_{\boldsymbol{\theta}}\left(\mathbf{X}\right)\right)\right]. 
\end{equation} The latter expectation still requires the computation of the intractable integral of $g_{\boldsymbol{\theta}}$ in \eqref{eq:g_theta_integral}. Nevertheless, by multiplying both sides of \eqref{eq:likelihood_lower_bound} by the positive weight $g^*\left(\mathbf{x}\right)/f\left(\mathbf{x}\right)$, we can obtain a computable lower bound of the weighted log-likelihood depending on the latent variable $\mathbf{z}$:\begin{equation}\label{eq:likelihood_lower_bound_weighed}
   \dfrac{g^*\left(\mathbf{x}\right)}{f\left(\mathbf{x}\right)}\log\left(g_{\boldsymbol{\theta}}\left(\mathbf{x}\right)\right) \geq \dfrac{g^*\left(\mathbf{x}\right)}{f\left(\mathbf{x}\right)}\left(\mathbb{E}_{q_{\boldsymbol{\phi}}\left(.\left|\mathbf{x}\right.\right)}\left[\log\left(g_{\boldsymbol{\theta}}\left(\mathbf{x}|\mathbf{Z}\right)\right)\right] - D_{\text{KL}}\left(q_{\boldsymbol{\phi}}\left(\mathbf{Z}|\mathbf{x}\right)\Vert p_{\boldsymbol{\lambda},\boldsymbol{\phi}}\left(\mathbf{Z}\right)\right)\right).
\end{equation} Finally, applying the expectation w.r.t. $\mathbf{X}\sim f$, we get:\begin{multline}\label{eq:elbo_weighed}
\mathbb{E}_{g^*}\left[\log\left(g_{\boldsymbol{\theta}}\left(\mathbf{X}\right)\right)\right] = \mathbb{E}_{f}\left[\dfrac{g^*\left(\mathbf{X}\right)}{f\left(\mathbf{X}\right)}\log\left(g_{\boldsymbol{\theta}}\left(\mathbf{X}\right)\right)\right]\\ \geq \mathbb{E}_{f}\left[\dfrac{g^*\left(\mathbf{X}\right)}{f\left(\mathbf{X}\right)}\left(\mathbb{E}_{q_{\boldsymbol{\phi}}\left(.|\mathbf{X}\right)}\left[\log\left(g_{\boldsymbol{\theta}}\left(\mathbf{X}|\mathbf{Z}\right)\right)\right] - D_{\text{KL}}\left(q_{\boldsymbol{\phi}}\left(\mathbf{Z}|\mathbf{X}\right)\Vert p_{\boldsymbol{\lambda},\boldsymbol{\phi}}\left(\mathbf{Z}\right)\right)\right)\right].
\end{multline} As in Section \ref{sec:vae_review}, the training procedure of the VAE will here aim to maximise the lower bound of \eqref{eq:elbo_weighed} over the data. At last, even though $f$ and $g^*$ are known only up to positive constants $c_f$ and $c_g$, one can ignore these constants in the expression of the function to maximise. Therefore, the objective function in the current framework is given by the new weighted ELBO function: \begin{equation}
\label{eq:wELBO}
    \text{wELBO}\left(\boldsymbol{\phi},\boldsymbol{\theta},\boldsymbol{\lambda}\right) =\mathbb{E}_{f}\left[\dfrac{\widetilde{g}^*\left(\mathbf{X}\right)}{\widetilde{f}\left(\mathbf{X}\right)}\mathbb{E}_{q_{\boldsymbol{\phi}}\left(.|\mathbf{X}\right)}\left[\log\left(g_{\boldsymbol{\theta}}\left(\mathbf{X}|\mathbf{Z}\right)\right)\right]\right] - \mathbb{E}_{f}\left[\dfrac{\widetilde{g}^*\left(\mathbf{X}\right)}{\widetilde{f}\left(\mathbf{X}\right)}D_{\text{KL}}\left(q_{\boldsymbol{\phi}}\left(\mathbf{Z}|\mathbf{X}\right)\Vert p_{\boldsymbol{\lambda},\boldsymbol{\phi}}\left(\mathbf{Z}\right)\right)\right].
\end{equation}

To sum up, only by modifying the objective function of the classical VAE from Section \ref{sec:vae_review}, it is possible to learn the high-dimensional target distribution $g^*$ from samples distributed according to an initial distribution $f$ with a VAE. In practice, all the terms of this objective function are estimated using the observed sample $\left(\mathbf{X}^{(n)}\right)_{n\in[\![1,N]\!]}$ distributed according to $f$.

\section{New initialisation procedure of the weights of the neural networks to prevent posterior collapse}
\label{sec:posterior_collapse}

\subsection{Posterior collapse}

A classical problem that badly affects the accuracy of the density estimation and the generating properties of the VAE is \textit{posterior collapse}. As described in many articles \cite{bowman2015generating,higgins2016beta,sonderby2016ladder,he2019lagging}, posterior collapse generally refers to an over-regularisation of the VAE, a loss of information, which mathematically traduces that the Kullback-Leibler term of the objective function vanishes, that is $D_{\text{KL}}\left(q_{\boldsymbol{\phi}}\left(\mathbf{Z}|\mathbf{x}\right)\Vert p_{\boldsymbol{\lambda},\boldsymbol{\phi}}\left(\mathbf{Z}\right)\right) \approx 0$ for every $\mathbf{x}\in\mathcal{X}$. In order words, every variational posterior distribution collapse into the prior. Getting stuck in a local maxima during the optimisation can explain this phenomenon \cite{sonderby2016ladder,lucas2019understanding}. Notably, a trained VAE affected by posterior collapse is not able to catch the different modes of a multimodal distribution. Note that \cite{takida2020ar} investigates an alternative definition of posterior collapse based on the mutual information.

The most classical solution to deal with posterior collapse is to introduce a coefficient $\beta\in[0,1]$ before the Kullback-Leibler term of the objective function in order to reduce the regularisation effect \cite{bowman2015generating,higgins2016beta,sonderby2016ladder}, but the corresponding objective function is no longer a lower bound of the log-likelihood of the model. Others papers propose an alternative formulation of the objective function \cite{rezende2018taming,alemi2018fixing} whereas the works of \cite{razavi2019preventing,xu2018spherical} investigate new choices for the family of distributions for the prior or the variational posterior.

\subsection{New pre-initialisation procedure}

In the present article, we introduce a new solution to handle posterior collapse. As explained above, posterior collapse is caused by getting stuck in a local maximum of the objective function. In order to prevent it, we propose a pre-initialisation procedure of the weights of the neural networks $\boldsymbol{\phi}$, $\boldsymbol{\theta}$ and $\boldsymbol{\lambda}$ in order to start the training of the VAE from more adapted starting points $\boldsymbol{\phi}^{(0)}$, $\boldsymbol{\theta}^{(0)}$ and $\boldsymbol{\lambda}^{(0)}$.

First, we initialise the weights $\boldsymbol{\lambda}$ by supervised learning. To do so, we randomly pick without replacement a sequence of indices $\left(s(k)\right)_{k\in[\![1,K]\!]}$ of integers of $[\![1,N]\!]$ with probabilities proportional to the family $\left(\widetilde{g}^*\left(\mathbf{X}^{(n)}\right)/\widetilde{f}\left(\mathbf{X}^{(n)}\right)\right)_{n\in[\![1,N]\!]}$ in order to create the sub-sample $\left(\mathbf{X}^{(s(k))}\right)_{k\in[\![1,K]\!]}$. Then, we train the neural network $\text{VP}_{\boldsymbol{\lambda}}$ such that it maps each vector $\mathbf{e}_k$ into the corresponding picked point from the dataset $\mathbf{X}^{(s(k))}$ by minimising the mean square error:
\begin{equation}\label{eq:loss_VPnet}
    \boldsymbol{\lambda^{(0)}} = \underset{\boldsymbol{\lambda}}{\arg\min} \sum_{k=1}^K \left(\text{VP}_{\boldsymbol{\lambda}}\left(\mathbf{e}_k\right) - \mathbf{X}^{(s(k))}\right)^\top\left(\text{VP}_{\boldsymbol{\lambda}}\left(\mathbf{e}_k\right) - \mathbf{X}^{(s(k))}\right).
\end{equation} The initial pseudo-inputs $\boldsymbol{u}_k^{(0)} = \text{VP}_{\boldsymbol{\lambda^{(0)}}}\left(\mathbf{e}_k\right)$ are thus already representative of the target distribution $g^*$.

Second, we initialise the weights $\boldsymbol{\phi}$ and $\boldsymbol{\theta}$ by unsupervised learning. To do so, in order to already ensure good reconstruction properties, we train the pair encoder/decoder $\left(E_{\boldsymbol{\phi}},D_{\boldsymbol{\theta}}\right)$ as a classical autoencoder on the whole dataset by minimising the mean square error between the data points and the reconstructed ones (encoded and then decoded). Once more, thanks to the importance sampling trick, we are able to rewrite the loss function as an expectation under $f$. Moreover, since a classical autoencoder only cares about the mean values $\boldsymbol{\mu}_{\mathbf{x}}$ and $\boldsymbol{\mu}_{\mathbf{z}}$, we add a penalisation term in order to also pre-train the weights of the networks corresponding to the scale parameters of the encoder in order to initialise the diagonal terms of $\boldsymbol{\Sigma}_{\mathbf{x}}$ close to $1$ (for normalized data). 

Writing $\left(E_{\boldsymbol{\phi}}^{\boldsymbol{\mu}}\left(\mathbf{x}\right),E_{\boldsymbol{\phi}}^{\boldsymbol{\Sigma}}\left(\mathbf{x}\right)\right) = E_{\boldsymbol{\phi}}\left(\mathbf{x}\right)$ and $\left(D_{\boldsymbol{\theta}}^{\boldsymbol{\mu}}\left(\mathbf{z}\right),D_{\boldsymbol{\theta}}^{\boldsymbol{\Sigma}}\left(\mathbf{z}\right)\right) = D_{\boldsymbol{\theta}}\left(\mathbf{z}\right)$ and letting $\log^2\left(D\right)$ be obtained by taking the square logarithm of a diagonal matrix D, we let: \begin{align}
    &\left(\boldsymbol{\phi}^{(0)},\boldsymbol{\theta}^{(0)}\right) = \underset{\boldsymbol{\phi},\boldsymbol{\theta}}{\arg\min} \ \mathbb{E}_{g^*}\left[\left(\mathbf{X} - D_{\boldsymbol{\theta}}^{\boldsymbol{\mu}}\left(E_{\boldsymbol{\phi}}^{\boldsymbol{\mu}}\left(\mathbf{X}\right)\right)\right)^\top\left(\mathbf{X}-D_{\boldsymbol{\theta}}^{\boldsymbol{\mu}}\left(E_{\boldsymbol{\phi}}^{\boldsymbol{\mu}}\left(\mathbf{X}\right)\right)\right) + \dfrac{1}{d_z}\mathrm{Tr}\left(\log^2\left(E_{\boldsymbol{\phi}}^{\boldsymbol{\Sigma}}\left(\mathbf{X}\right)\right)\right)\right] \notag\\ &= \underset{\boldsymbol{\phi},\boldsymbol{\theta}}{\arg\min} \ \mathbb{E}_{f}\left[\dfrac{g^*\left(\mathbf{X}\right)}{f\left(\mathbf{X}\right)}\left(\left(\mathbf{X}-D_{\boldsymbol{\theta}}^{\boldsymbol{\mu}}\left(E_{\boldsymbol{\phi}}^{\boldsymbol{\mu}}\left(\mathbf{X}\right)\right)\right)^\top\left(\mathbf{X}-D_{\boldsymbol{\theta}}^{\boldsymbol{\mu}}\left(E_{\boldsymbol{\phi}}^{\boldsymbol{\mu}}\left(\mathbf{X}\right)\right)\right)+ \dfrac{1}{d_z}\mathrm{Tr}\left(\log^2\left(E_{\boldsymbol{\phi}}^{\boldsymbol{\Sigma}}\left(\mathbf{X}\right)\right)\right)\right)\right] \notag\\ &= \underset{\boldsymbol{\phi},\boldsymbol{\theta}}{\arg\min} \ \mathbb{E}_{f}\left[\dfrac{\widetilde{g}^*\left(\mathbf{X}\right)}{\widetilde{f}\left(\mathbf{X}\right)}\left(\left(\mathbf{X}-D_{\boldsymbol{\theta}}^{\boldsymbol{\mu}}\left(E_{\boldsymbol{\phi}}^{\boldsymbol{\mu}}\left(\mathbf{X}\right)\right)\right)^\top\left(\mathbf{X}-D_{\boldsymbol{\theta}}^{\boldsymbol{\mu}}\left(E_{\boldsymbol{\phi}}^{\boldsymbol{\mu}}\left(\mathbf{X}\right)\right)\right)+ \dfrac{1}{d_z}\mathrm{Tr}\left(\log^2\left(E_{\boldsymbol{\phi}}^{\boldsymbol{\Sigma}}\left(\mathbf{X}\right)\right)\right)\right)\right]. \label{eq:loss_autoencoder}
\end{align} 

At last, the full proposed training procedure of the VAE with weighted samples is given in Algorithm \ref{algo:vae_is}. Note that this procedure can easily be adapted to the classical VAE framework described in Section \ref{sec:vae_review} only by removing the likelihood ratio $\widetilde{g}^*\left(\mathbf{x}\right) \left/ \widetilde{f}\left(\mathbf{x}\right) \right.$.  \begin{algorithm}
\caption{VAE-IS}
\begin{algorithmic}[1]
\REQUIRE $\left(\mathbf{X}^{(n)}\right)_{n\in[\![1,N]\!]}\sim f$,$K$,$d_{\boldsymbol{z}}$
\STATE Randomly pick $K$ points $\left(\mathbf{X}^{(s(k))}\right)_{k\in[\![1,K]\!]}$ within the dataset 
\STATE Train $\text{VP}_{\boldsymbol{\lambda}}$ by minimising \eqref{eq:loss_VPnet} and get $\boldsymbol{\lambda}^{(0)}$
\STATE Train $\left(E_{\boldsymbol{\phi}},D_{\boldsymbol{\theta}}\right)$ by minimising \eqref{eq:loss_autoencoder} and get $\left(\boldsymbol{\phi}^{(0)},\boldsymbol{\theta}^{(0)}\right)$
\STATE Train the whole VAE $\left(E_{\boldsymbol{\phi}},D_{\boldsymbol{\theta}},\text{VP}_{\boldsymbol{\lambda}}\right)$ by maximising \eqref{eq:wELBO} starting from $\left(\boldsymbol{\phi}^{(0)},\boldsymbol{\theta}^{(0)},\boldsymbol{\lambda}^{(0)}\right)$
\RETURN Trained VAE.
\end{algorithmic}
\label{algo:vae_is}
\end{algorithm}

\begin{remark}\label{rk:comparison}
The objective of \cite{wang2019adaptive} is the same as ours: learning a target distribution $g^*$ by using observations from an initial distribution $f$ with a VAE. The main principle of their algorithm is the following: \begin{enumerate}
    \item train a classical VAE as in Section \ref{sec:vae_review} with the available observations from $f$,
    \item draw new points from the trained VAE and compute the corresponding likelihood ratios,
    \item boostrap among the new sample with probabilities proportional to the likelihood ratios.
\end{enumerate} They iteratively repeat this procedure and the final sample is theoretically distributed according to a distribution close to $g^*$. Contrary to their work, we propose a one-step procedure to achieve the same goal by modifying the objective function of the VAE. Moreover, it does not seem clear how to have access to the PDF of the resulting approximating distribution in \cite{wang2019adaptive} (if at all possible) whereas it is straightforward with our approach. In addition, we also introduce a new pre-training procedure of the weights of the neural networks in order to prevent posterior collapse. At last, they only test their algorithm on low dimensional (10 and 6) uni-modal target distributions whereas, as developed in Section \ref{sec:numerical_results}, we test our procedure on more complex test cases.
\end{remark}

\FloatBarrier

\section{Application to high-dimensional non-parametric importance sampling}
\label{sec:numerical_results}
In this section, we first show how to apply the proposed procedure to the classical framework of importance sampling. Then, in order to illustrate the practical interest of the previous efforts, we evaluate numerically the performances of the suggested procedure and we compare them to the performances of some existing IS methods. The code to reproduce the numerical experiments is publicly available at: \url{https://github.com/Julien6431/Importance-Sampling-VAE.git}.

\subsection{Importance sampling in high dimension}
\label{ssec:is_pres}

\subsubsection{General presentation of importance sampling}

Importance sampling (IS) is a classical variance-reduction technique which was introduced in \cite{Kahn1951SplittingParticleTransmission} and massively used for reliability analysis \cite{Shinozuka1983BasicAO,papaioannou2019improved,chiron2023failure} in particular. In the case of the estimation of an expectation $I = \mathbb{E}_{f}\left(\psi\left(\mathbf{X}\right)\right)$ where $\psi : \mathcal{X}\longrightarrow \mathbb{R}$ is a black-box function, it consists in rewriting the expectation according to an auxiliary density $g : \mathcal{X} \longrightarrow \mathbb{R}_+$  as $\mathbb{E}_{g}\left(\psi\left(\mathbf{X}\right)w^g\left(\mathbf{X}\right)\right)$, where $w^g\left(\mathbf{x}\right) = f(\mathbf{x})/g(\mathbf{x})$ is the likelihood ratio. To get an unbiased estimate, the support of $g$ must contain the support of $\mathbf{x}\in\mathcal{X} \mapsto \psi\left(\mathbf{x}\right)f(\mathbf{x)}$. The corresponding estimator is then given by: \begin{equation}\label{eq:ptisest}\widehat{I}_{g,N}^{\text{IS}} = \frac{1}{N}\sum_{n=1}^N \psi\left(\mathbf{X}^{(n)}\right)w^g\left(\mathbf{X}^{(n)}\right),\end{equation} where $\left(\mathbf{X}^{(n)}\right)_{n\in [\![1,N]\!]}$ is an i.i.d. sample distributed according to the IS auxiliary distribution $g$. It is consistent and unbiased, and it has zero-variance if and only if $g = g_{\text{opt}}$ with $\forall \mathbf{x} \in \mathbb{X}$, $g_{\text{opt}}\left(\mathbf{x}\right) \propto \psi\left(\mathbf{x}\right)f(\mathbf{x)}$ \cite{bucklew2004introduction} on the condition that $\psi$ is non-negative. This optimal density cannot be used in practice because the normalizing constant is $I$, which is the quantity to estimate.

In practice, the optimal IS distribution $g_{\text{opt}}$ is approximated non-parametrically \cite{zhang2006bayesian,morio2011non} or within a parametric family of distributions. As explained at the beginning of Section \ref{sec:vae_review}, non-parametric methods strongly suffer from the curse of dimensionality. The most common and convenient parametric auxiliary families are the \textit{Gaussian} family \cite{rubinstein2004cross,de2005tutorial} and the \textit{Gaussian mixture} family \cite{kurtz2013cross,geyer2019cross} for multimodal target distribution $g_{\text{opt}}$. However, the amount of parameters to estimate for these distributions makes them inaccurate when the dimension $d$ increases. In the specific case where the input distribution $f$ is the standard Gaussian distribution, the \textit{von Mises–Fisher–Nakagami} (vMFNM) family of distributions \cite{mardia2000directional,wang2016cross,papaioannou2019improved} is well-suited and more robust when the dimension $d$ is high. However, when considering mixtures of vMFNM distributions for multimodal target distributions, the corresponding algorithms require the knowledge of the number of modes. 

\subsubsection{Importance sampling supported by a VAE}

In this paper, we propose to select the IS auxiliary distribution within the distributions parameterised by VAEs. In the IS framework, the target distribution corresponding to $g^*$ of Section \ref{sec:new_vae_procedure} is here $g_{\text{opt}}$, which is known up to constant. Moreover, we assume that the initial sampling distribution $f$ is perfectly known. Then, in order to compute the likelihood ratios $w^g$ of the estimator in \eqref{eq:ptisest}, let us explain how to get the corresponding PDF of the resulting distribution $g_{\boldsymbol{\theta}}$. The most naive way to do so is \cite{wang2019adaptive}: \begin{enumerate}
    \item draw a sample $\left(\mathbf{X}^{(n)},\mathbf{Z}^{(n)}\right)_{n\in[\![1,N]\!]}$ according to the joint distribution $g_{\boldsymbol{\theta}}\left(\mathbf{x},\mathbf{z}\right) = p\left(\mathbf{z}\right)g_{\boldsymbol{\theta}}\left(\mathbf{x}\left|\mathbf{z}\right.\right)$ on $\mathcal{X}\times\mathcal{Z}$,
    \item estimate the PDF values associated to the sample with: \begin{equation}
    \widehat{g_{\boldsymbol{\theta}}\left(\mathbf{X}^{(n)}\right)} = \dfrac{1}{N}\sum_{k=1}^{N}g_{\boldsymbol{\theta}}\left(\mathbf{X}^{(n)}\left|\mathbf{Z}^{(k)}\right.\right).
\end{equation}
\end{enumerate} However, the use of these estimated values of the PDF makes the IS estimator in \eqref{eq:ptisest} biased since $\widehat{g_{\boldsymbol{\theta}}\left(\mathbf{X}^{(n)}\right)}$ is at the denominator. Moreover, the PDF values $\left(\widehat{g_{\boldsymbol{\theta}}\left(\mathbf{X}^{(n)}\right)}\right)_{n\in[\![1,N]\!]}$ are not independent. Thus, the estimator in \eqref{eq:ptisest} is no longer a sum of independent random variables and as a result, its convergence is no longer guaranteed.

Hence, in order to keep the convenient properties of the classical IS estimator in \eqref{eq:ptisest}, we adopt the following procedure: \begin{enumerate}
    \item approximate the marginal distribution $g_{\boldsymbol{\theta}}$ by:\begin{equation}
        \forall\mathbf{x}\in\mathcal{X}, \ g_{\boldsymbol{\theta}}^M\left(\mathbf{x}\right) = \dfrac{1}{M}\sum_{m=1}^Mg_{\boldsymbol{\theta}}\left(\mathbf{x}\left|\mathbf{Z}^{(m)}\right.\right),
    \end{equation}where $M\geq 1$ and where $\left(\mathbf{Z}^{(m)}\right)_{m\in[\![1,M]\!]}$ is an i.i.d. sample from the latent space $\mathcal{Z}$ distributed according to the prior distribution $p$,
    \item draw a sample $\left(\mathbf{X}^{(n)}\right)_{n\in[\![1,N]\!]}$ according to $g_{\boldsymbol{\theta}}^M$ and compute the corresponding PDF values $g_{\boldsymbol{\theta}}^M\left(\mathbf{X}^{(n)}\right)$ for all $n\in[\![1,N]\!]$.
\end{enumerate} The distribution $g_{\boldsymbol{\theta}}^M$ is an approximation of $g_{\boldsymbol{\theta}}$ but allows to compute the exact PDF values associated to the generated points and then to keep the convenient statistical properties of the IS estimator. Also, conditionally to $\left(\mathbf{Z}^{(m)}\right)_{m\in[\![1,M]\!]}$, $\left(w^{g_{\boldsymbol{\theta}}}\left(\mathbf{X}^{(n)}\right)\right)_{n\in[\![1,N]\!]}$ remain independent.

\begin{figure}
    \centering
    \includegraphics[width=\textwidth]{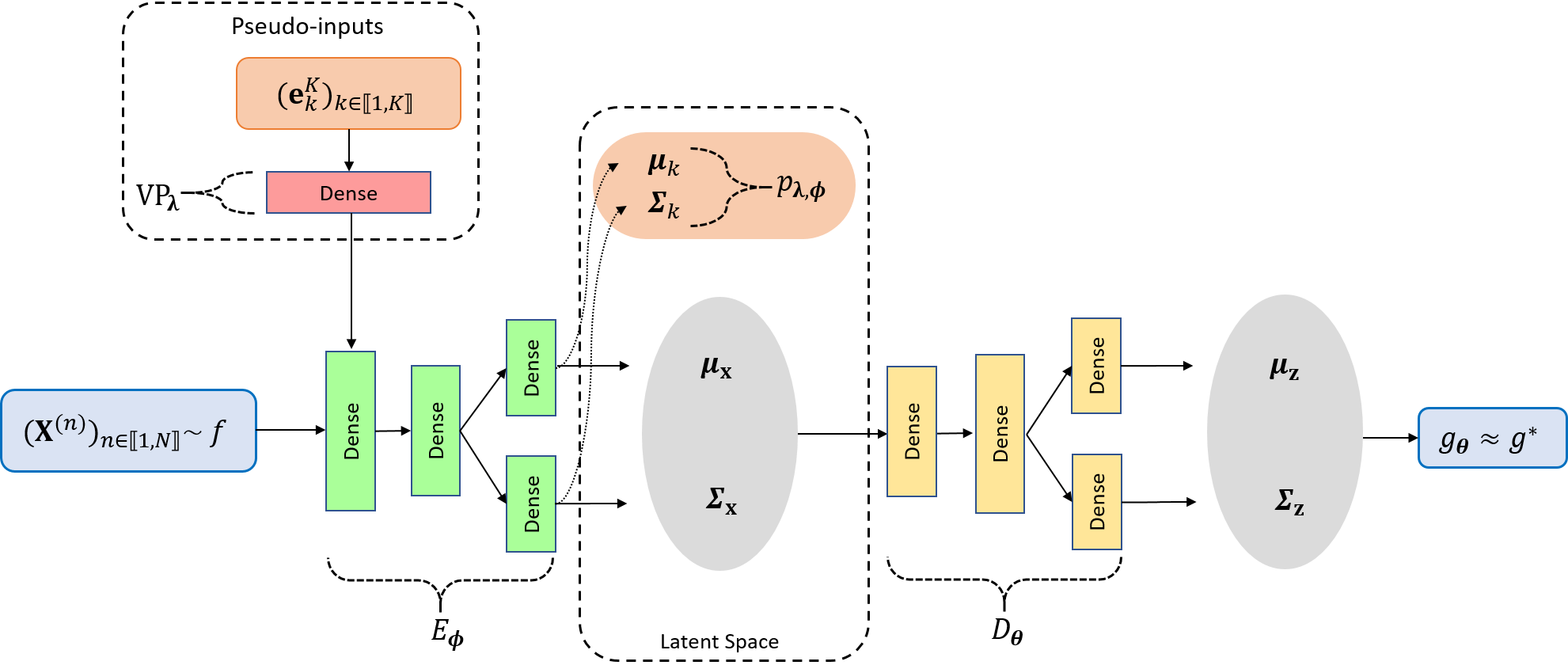}
    \caption{Representation of our suggested VAE architecture for IS in high dimension.}
    \label{fig:vae_archi}
\end{figure}

The detailed architecture of the VAE used for the following test cases is shown in Figure \ref{fig:vae_archi}. Moreover, for each test case, we set $K=75$ and $M=10^3$, and we will explicit the dimension $d_z$ of the latent space chosen for each example.

\subsection{Generation of samples by adaptive IS}

A first application of density estimation with weighted samples is adaptive IS for sampling from a target distribution $g^*$. Sometimes, sampling directly from the target distribution $g^*$ is not possible, especially when it is only known up to a constant \cite{ghosal2017fundamentals}. Then, starting from an initial proposal distribution $f$, adaptive IS algorithms consist in iteratively approximating the target distribution within a family of proposal distributions and at the end, the resulting sample is expected to be drawn according to a distribution close to $g^*$.

In both examples, we use a classical adaptive IS scheme (AIS) as in \cite{marin2019mamis}, except that we return the weighted sample generated only at the last iteration. We use a distribution parameterised by a VAE as the proposal distribution at each iteration (AIS-VAE).

\subsubsection{Example in dimension 10}

First, let us consider a $10$-dimensional bimodal target distribution given by: \begin{equation}
    g_1^* \propto \mathcal{N}\left(2.5\times\mathbf{1}_{10},\mathbf{I}_{10}\right) + \mathcal{N}\left(-2.5\times\mathbf{1}_{10},\mathbf{I}_{10}\right),
\end{equation} where $\mathbf{1}_{10} = \left(1,\dots,1\right)^\top \in \mathbb{R}^{10}$ and $\mathbf{I}_{10}$ is the $10$-dimensional identity matrix. For comparison purposes, we execute as well the AIS algorithm with a Gaussian mixture with two components (AIS-GM) as the proposal distribution. The standard Gaussian distribution in dimension 10 is the starting proposal distribution for both algorithms. We perform $n_{\text{rep}}=100$ executions of each algorithm with 10 iterations. Then, at each iteration, we draw $N=10^4$ new points, and the dimension of the latent space chosen for this problem is $d_z = 4$.

\begin{figure}
        \centering
        \begin{subfigure}[b]{0.475\textwidth}
            \centering
            \includegraphics[width=\textwidth]{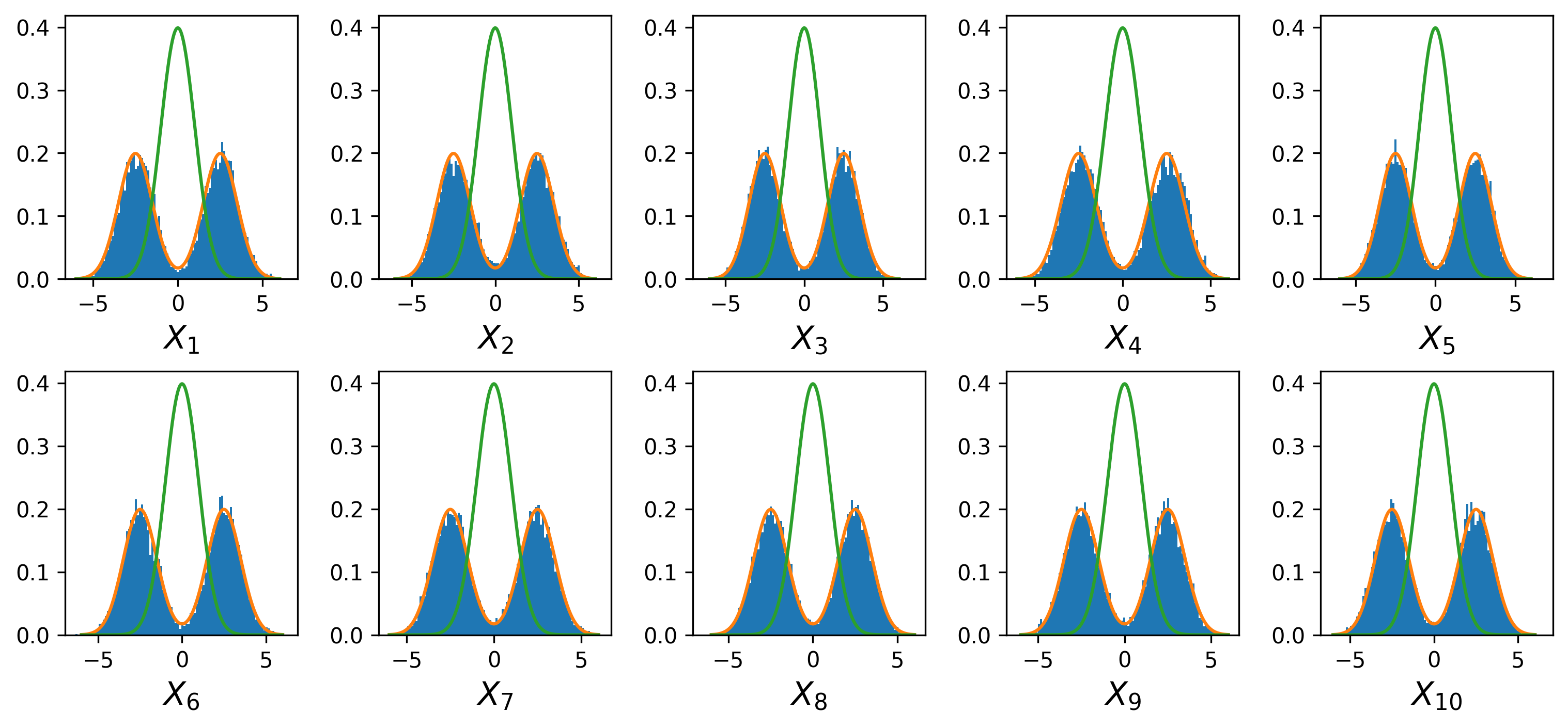}
            \caption{{\small Example for the AIS-VAE algorithm with 2 modes found.}} 
            \label{fig:AIS_vae2}
        \end{subfigure}
        \hfill
        \begin{subfigure}[b]{0.475\textwidth}  
            \centering 
            \includegraphics[width=\textwidth]{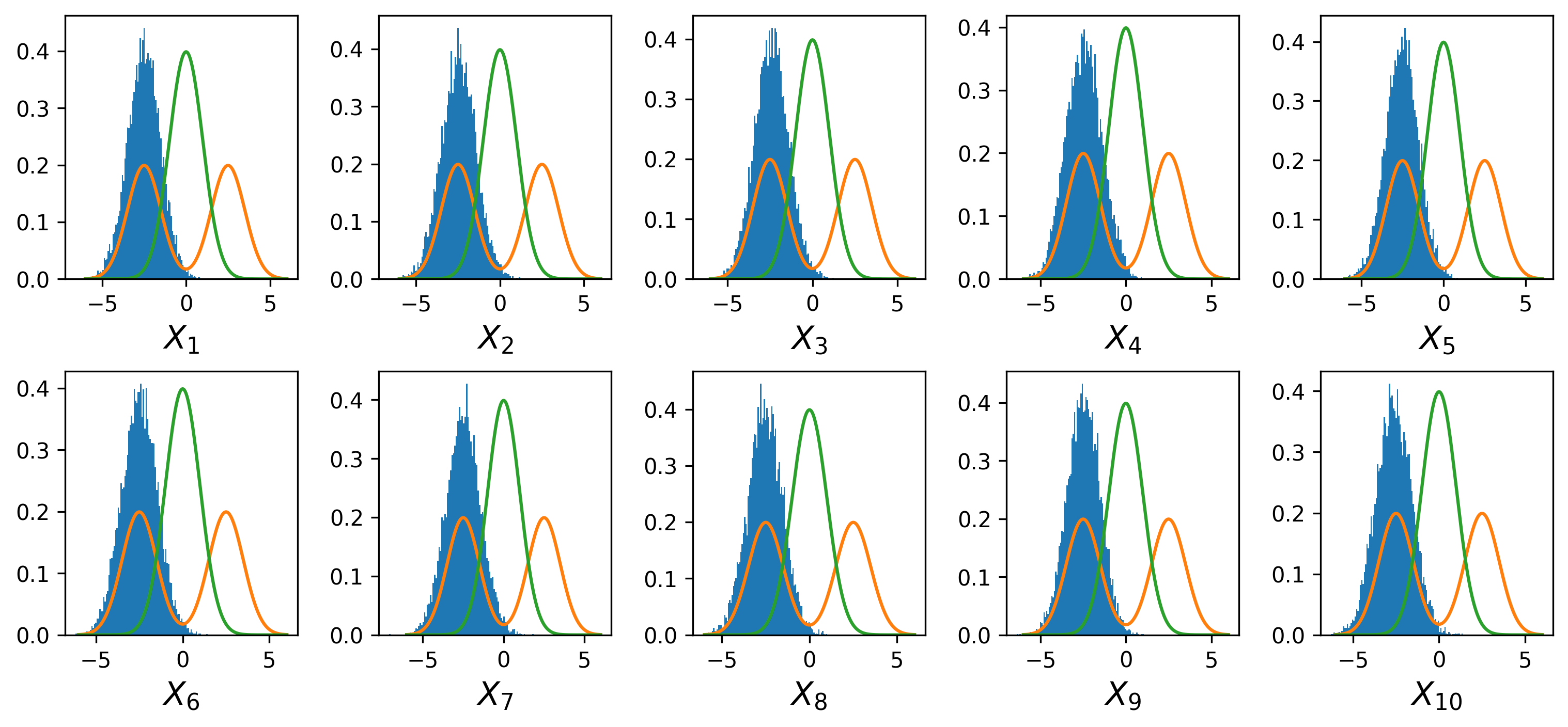}
            \caption{{\small Example for the AIS-VAE algorithm with 1 mode found.}}    
            \label{fig:AIS_vae1}
        \end{subfigure}
        \vskip\baselineskip
        \begin{subfigure}[b]{0.475\textwidth}   
            \centering 
            \includegraphics[width=\textwidth]{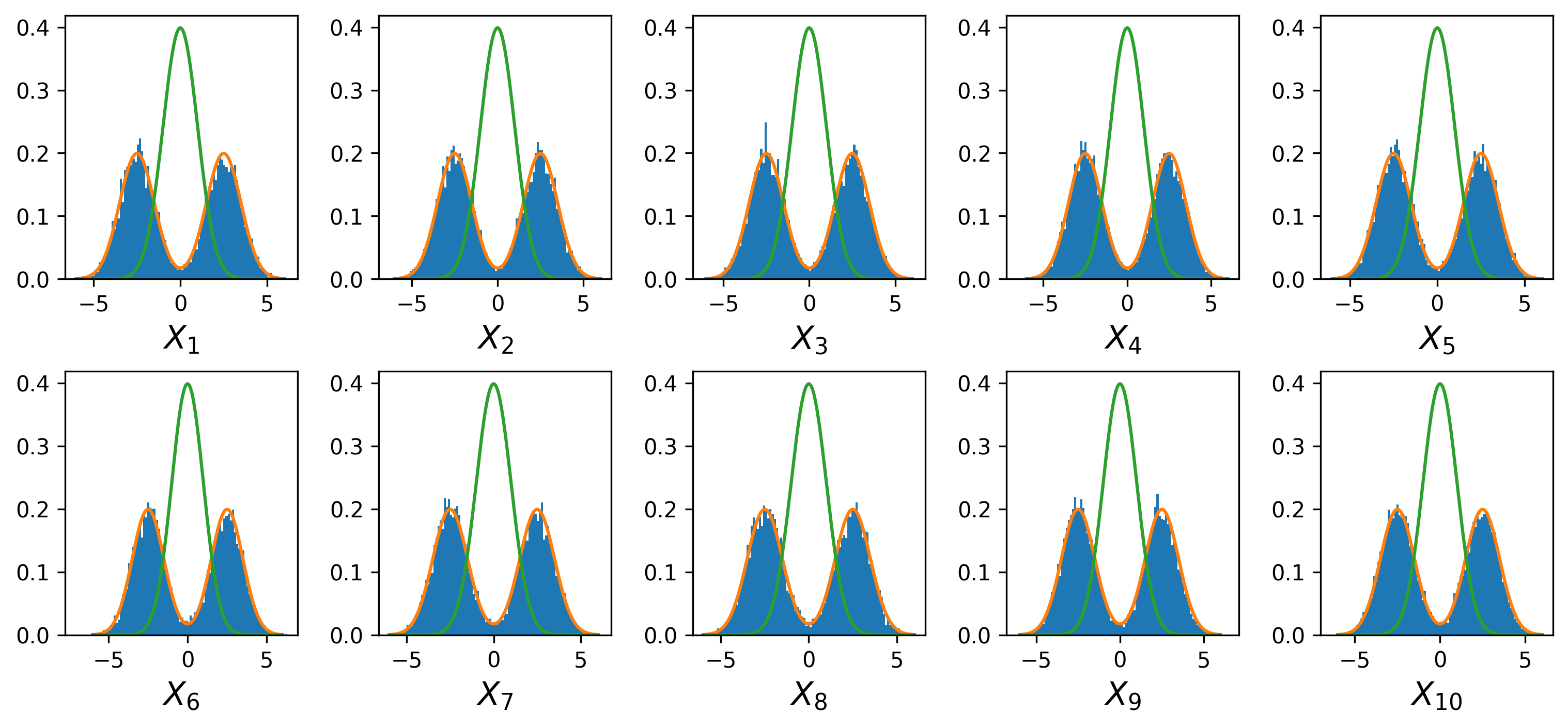}
            \caption{{\small Example for the AIS-GM algorithm with 2 modes found.}}    
            \label{fig:AIS_gm2}
        \end{subfigure}
        \hfill
        \begin{subfigure}[b]{0.475\textwidth}   
            \centering 
            \includegraphics[width=\textwidth]{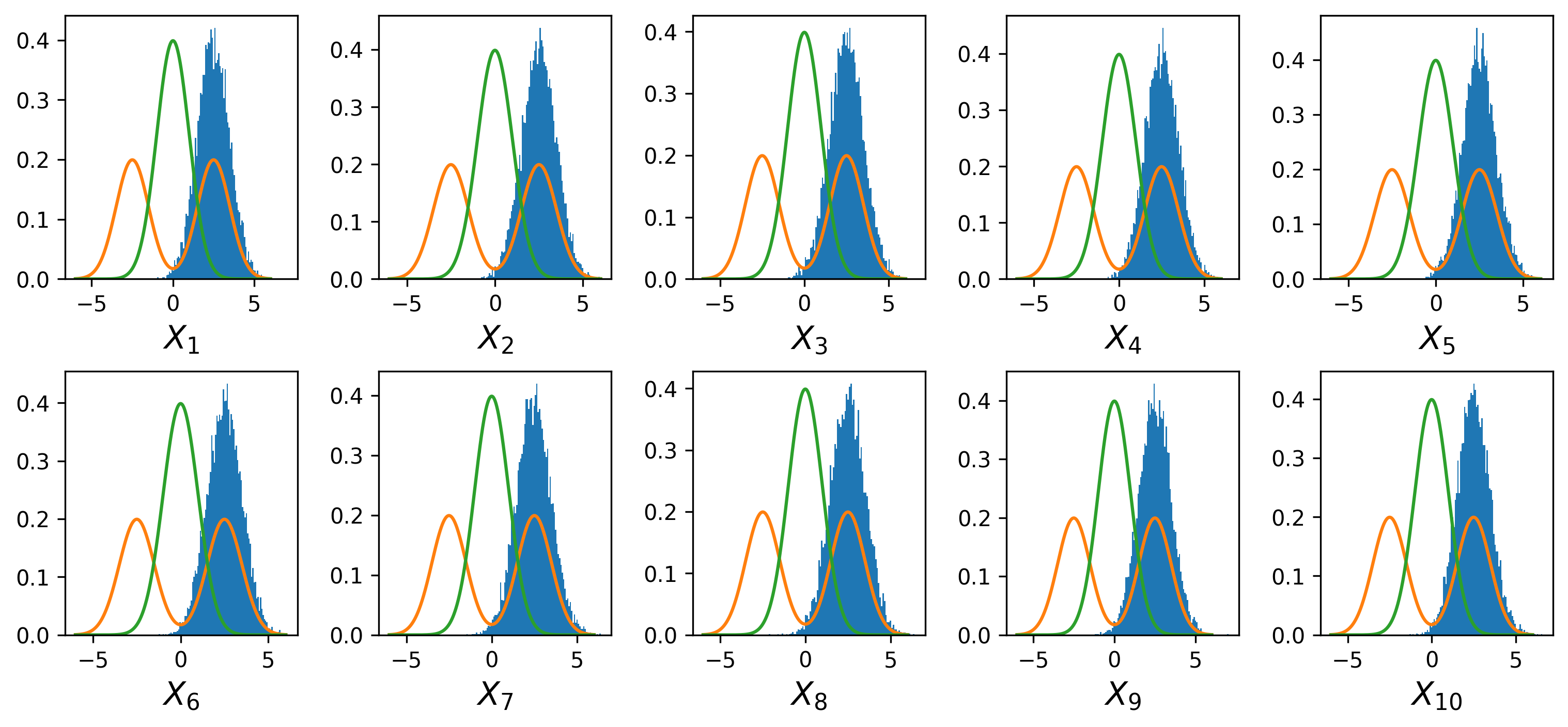}
            \caption{{\small Example for the AIS-GM algorithm with 1 mode found.}}    
            \label{fig:AIS_gm1}
        \end{subfigure}
        \caption{{\small Graphical representation as 10 histograms of the 10 marginals of the final generated sample for each case for both algorithms. The continuous orange line represents the PDF of the marginals of the target distribution $g^*_1$. The continuous green line represents the PDF of the marginals of the starting distribution $f$.} }
        \label{fig:generation_10}
\end{figure}

For both algorithms, the results are as follows: either both modes, and so the whole target distribution, are perfectly found and approximated, or only one mode is found but it is well approximated. Examples of the representation of the final approximating sample are given in Figure \ref{fig:generation_10}. However, the main difference between both algorithms is their success rate, i.e. the frequency with which the resulting distribution finds both modes. Indeed, as shown in Table \ref{table:generation_10}, the AIS-VAE algorithm finds both modes more than $70\%$ of the time and gives a very good approximation of the target distribution whereas the AIS-GM algorithm only finds both modes less than $10\%$ of the time over the $n_{\text{rep}}=100$ repetitions. These results show that a distribution parameterised by a VAE is more likely to catch the characteristics of a target distribution than a single Gaussian or a Gaussian mixture, and even though the latter knew the number of modes in advance. 

The likelihood ratios corresponding to the samples adaptively guide the algorithm from the initial distribution $f$ to the target one $g_1^*$. However, even more so when the dimension increases, one or some values can be significantly larger than the others, and as a result the adaptive IS algorithm will go in that direction. Nevertheless, we can not explain why this phenomenon seems such less important when using a VAE rather than a Gaussian mixture. A first possibility is that the dimensionality reduction performed by the VAE makes the process more robust. A second one can be the differences between the two learning procedures themselves. The learning procedure is based on mini-batches for the VAE whereas the whole dataset is used in a single step for learning the Gaussian mixture.

\begin{table}[h!]
\caption{Comparison of the AIS-VAE and AIS-GM algorithms. The first row of the table represents the success rate of the corresponding algorithm and the quantity $\left(D_{\text{KL}}\right)_{\text{mean}}^{\text{success}}$ represents the mean value of the Kullback-Leibler divergence between the empirical generated distribution and the target $g_1^*$ over the successful samples.}
\centering
\begin{tabular}{|c||c|c|} 
 \hline
& AIS-VAE & AIS-GM\\
 \hline\hline
Success rate & $72\%$  & $7\%$  \\ 
\hline
$\left(D_{\text{KL}}\right)_{\text{mean}}^{\text{success}}$ & $2.48\times 10^{-2}$ & $1.13\times 10^{-1}$  \\ 
\hline
\end{tabular}
\label{table:generation_10}
\end{table} 

\subsubsection{Example in dimension 20}

\begin{figure}
        \centering
        \begin{subfigure}[b]{0.475\textwidth}
            \centering
            \includegraphics[width=\textwidth]{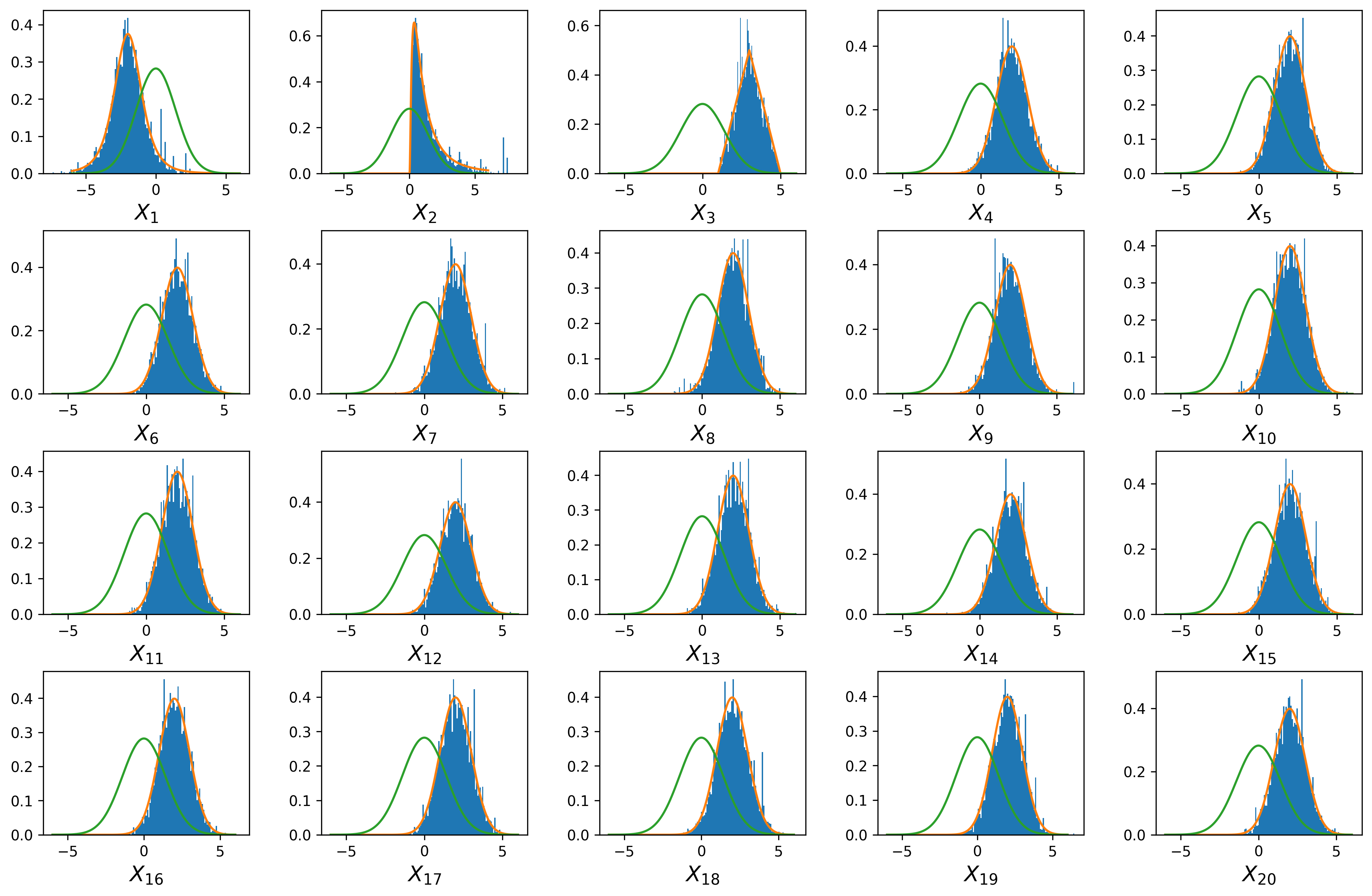}
            \caption{{\small }} 
            \label{fig:vae_20_0}
        \end{subfigure}
        \hfill
        \begin{subfigure}[b]{0.475\textwidth}  
            \centering 
            \includegraphics[width=\textwidth]{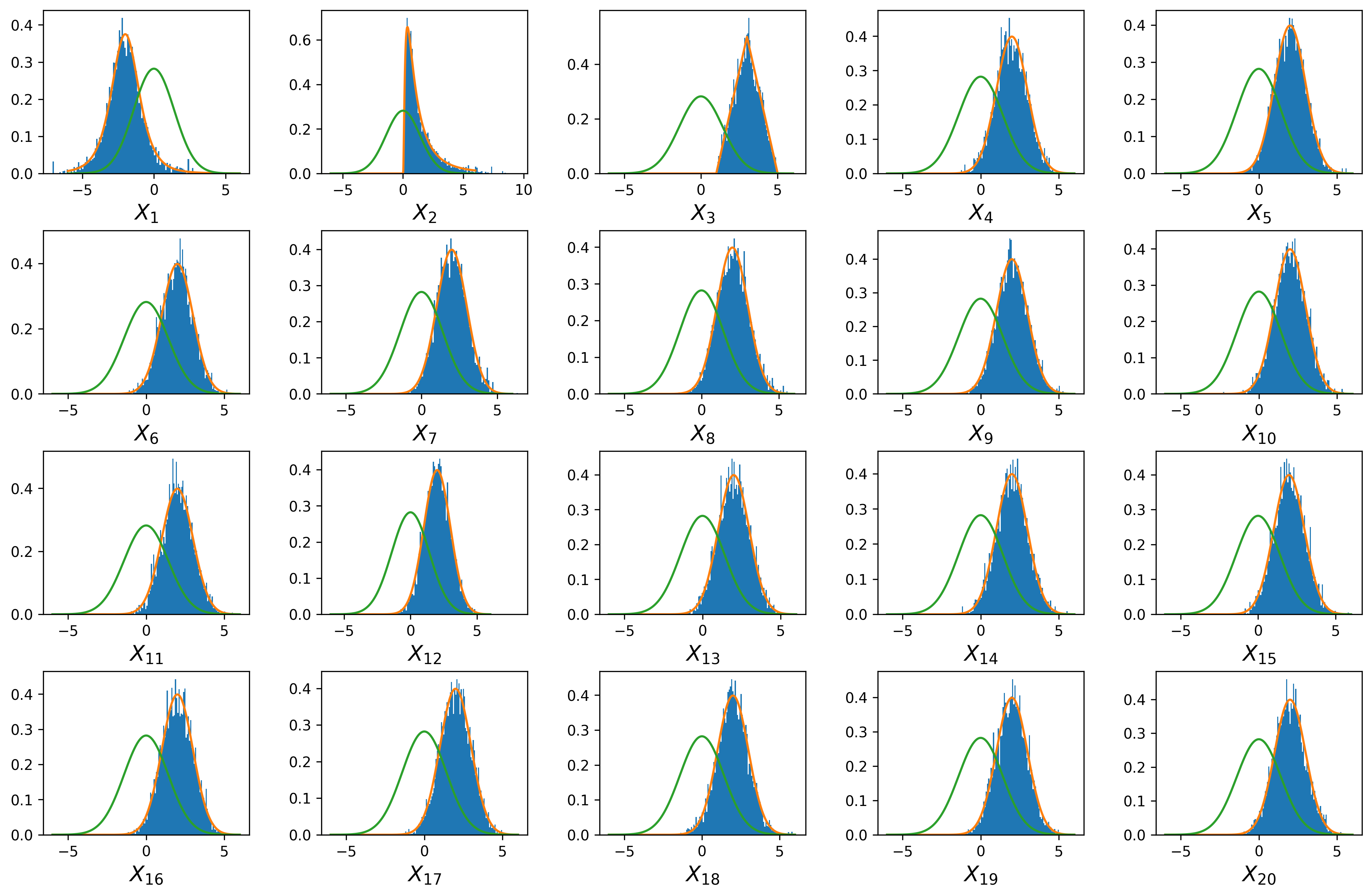}
            \caption{{\small }}
            \label{fig:vae_20_1}
        \end{subfigure}
        \vskip\baselineskip
        \begin{subfigure}[b]{0.475\textwidth}   
            \centering 
            \includegraphics[width=\textwidth]{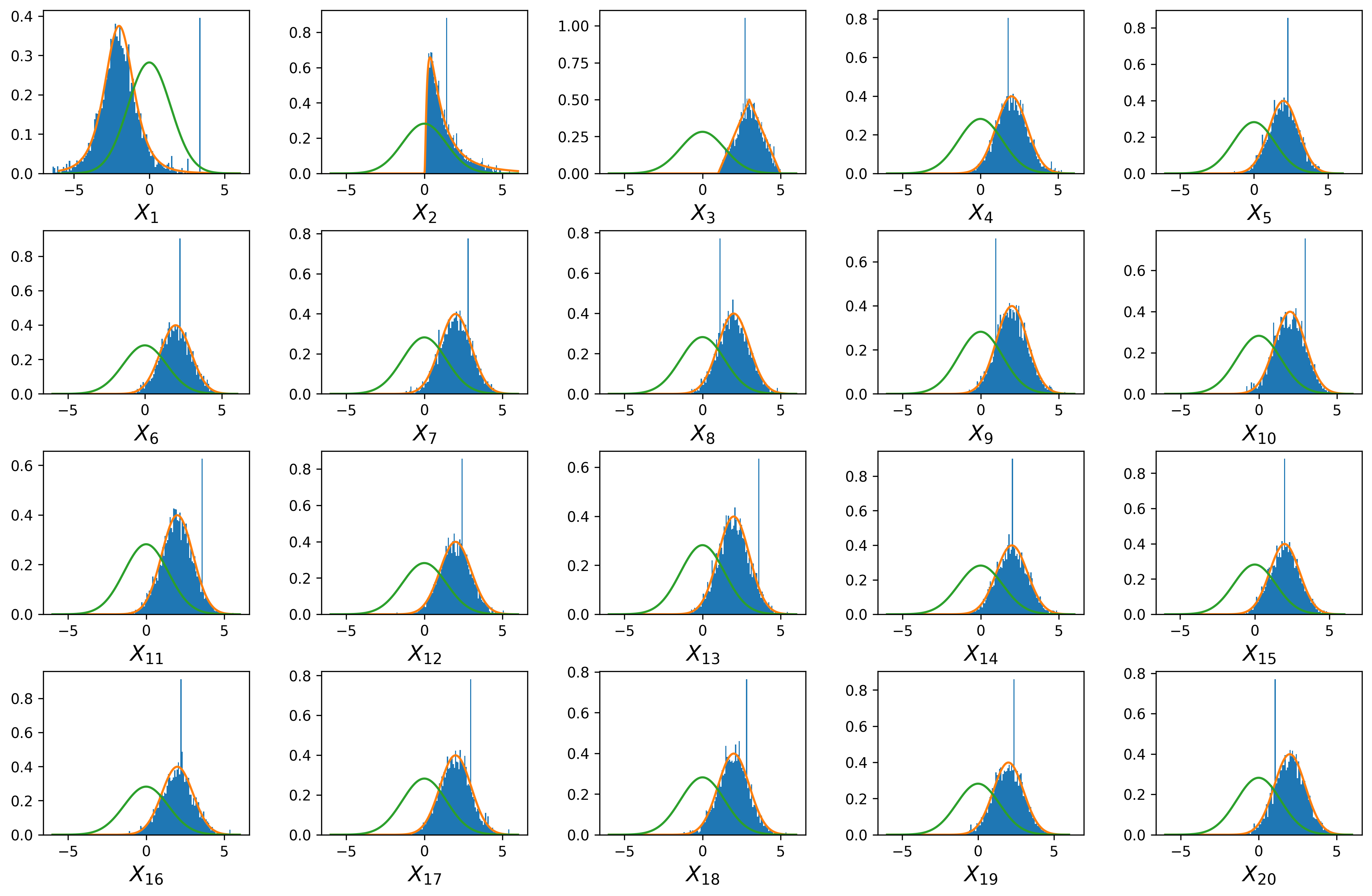}
            \caption{{\small}}
            \label{fig:vae_20_2}
        \end{subfigure}
        \hfill
        \begin{subfigure}[b]{0.475\textwidth}   
            \centering 
            \includegraphics[width=\textwidth]{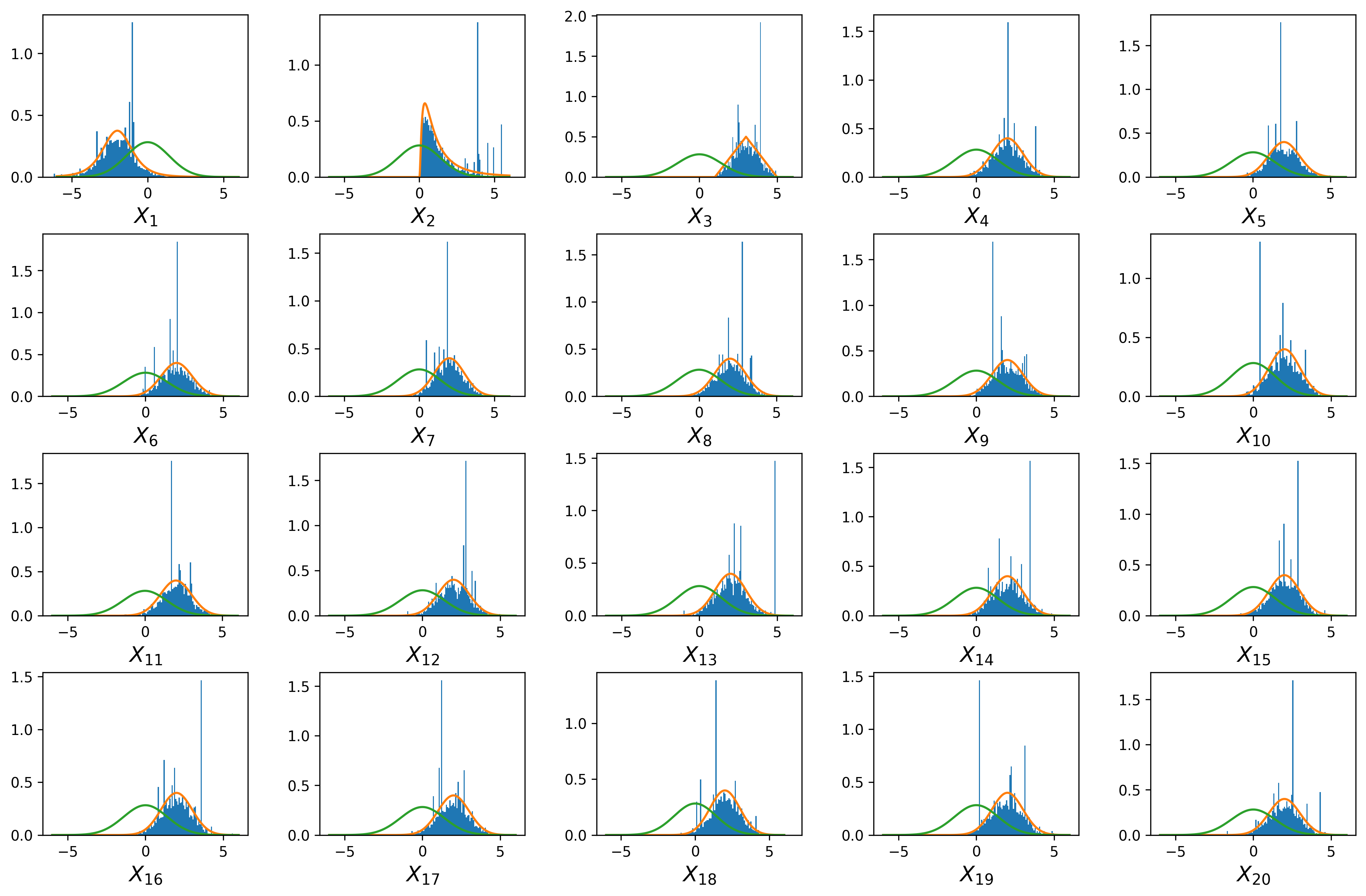}
            \caption{{\small }}  
            \label{fig:vae_20_3}
        \end{subfigure}
        \caption{{\small Graphical representation as 20 histograms of the 20 marginals of four realisations of the final generated sample for the AIS-VAE algorithm. The continuous orange line represents the PDF of the marginals of the target distribution $g^*_2$. The continuous green line represents the PDF of the marginals of the starting distribution $f$.} }
        \label{fig:generation_20}
\end{figure}

Second, let us consider a $20$-dimensional target distribution $g_2^*$ defined by the marginal distributions and the dependence structure given in Table \ref{table:target_g2}. The starting distribution chosen here is $f=\mathcal{N}\left(\boldsymbol{0}_{20},2\times \mathbf{I}_{20}\right)$. We perform 10 iterations and we draw $N=10^4$ new points at each one. The dimension of the latent space for this problem is $d_z = 8$. Examples of the resulting empirical distribution returned by the AIS-VAE algorithm can be found in Figure \ref{fig:generation_20}. We can see graphically that the target distribution is most of the time well approximated by the resulting distribution. However, because of very high values of the PDF, it is not possible to compute numerically the Kullback-Leibler divergence between the empirical distribution and the target distribution $g_2^*$, and then to use it as a quantitative measure of the quality of the approximation.

\begin{table}[h!]
\caption{$\text{Student}\left(\nu,\mu,\sigma\right)$: $1$-dimensional Student distribution with $\nu >0$ degrees of freedom, with mean $\mu\in \mathbb{R}$ and scale parameter $\sigma>0$. $\text{LogN}\left(\mu,\sigma\right)$: distribution of $\exp\left(A\right)$ with $A$ a $1$-dimensional Gaussian random variable of mean $\mu\in \mathbb{R}$ and scale $\sigma>0$. $\text{Triangular}\left(a,m,b\right)$: $1$-dimensional Triangular distribution where $a<b$ are the lower and upper bounds and where $m \in [a,b]$ is the mode. The correlation matrix is given by $R_{i,j} = \mathbf{1}\left(i=j\right) + 1/4\times \mathbf{1}\left(\left|i-j\right|=1\right)$ for all $\left(i,j\right)\in[\![1,20]\!]^2$.}
\centering
\begin{tabular}{|c||c|} 
 \hline
Input & Distribution\\
 \hline\hline
$X_1$  & $\text{Student}\left(4,-2,1\right)$  \\ 
\hline
$X_2$  & $\text{LogN}\left(0,1\right)$  \\ 
\hline
$X_3$  & $\text{Triangular}\left(1,3,5\right)$  \\ 
\hline
$X_4$ to $X_{20}$  & $\mathcal{N}\left(2,1\right)$  \\ 
\hline
Copula & Normal copula with correlation matrix $\mathbf{R}\in\mathbb{R}^{20\times 20}$ \\ 
\hline
\end{tabular}
\label{table:target_g2}
\end{table}

Moreover, for comparison purposes, we applied the AIS algorithm with a single Gaussian (AIS-SG) as the proposal distribution, since $g_1^*$ has only one mode. However, it does not work at all with the same setting. Indeed, the starting distribution $f = \mathcal{N}\left(\boldsymbol{0}_{20},2\times \mathbf{I}_{20}\right)$ is too far away from the target $g_2^*$ for the AIS-SG algorithm. Then, it is necessary to choose $f$ closer to $g_2^*$, by modifying the mean vector for example. Once more, this example shows that using a distribution parameterised by a VAE allows to have a larger flexibility than with a classical single Gaussian distribution, because the corresponding adaptive IS algorithms have a larger range.

Nevertheless, because of the weight degeneracy phenomenon \cite{rubinstein2009deal} which occurs in IS when the dimension increases, the AIS-VAE algorithm, in the same manner as the other adaptive IS algorithms, will not work very well in higher dimension than 20.

\subsection{Adaptive importance sampling for estimating failure probabilities}

Another application of density estimation with weighted samples is adaptive IS for reliability analysis. Reliability analysis consists in the estimation of the failure probability $p_t = \mathbb{P}\left(\psi\left(\mathbf{X}\right)>t\right) = \mathbb{E}_f\left[\mathbf{1}\left(\psi\left(\mathbf{X}\right)>t\right)\right]$ for a fixed known threshold $t\in\mathbb{R}$. Classical Monte Carlo sampling is not adapted to this problem when $p_t$ is getting smaller because its computational cost becomes too large to obtain an accurate estimation. Therefore, several techniques have been developed in order to reduce the variance of the estimation of $p_t$: one can mention FORM/SORM methods \cite{hasofer1974exact,breitung1984asymptotic}, subset sampling \cite{cerou:inria-00584352} or line sampling \cite{koutsourelakis2004reliability} for example.

In the IS framework, because of the rarity of the failure event, it is challenging to directly pick the best representative of $g_{\text{opt}}\left(\mathbf{x}\right)\propto\mathbf{1}\left(\psi\left(\mathbf{x}\right)>t\right)f\left(\mathbf{x}\right)$ within the parametric family of densities by minimising the KL divergence to $g_{\text{opt}}$. To overcome this issue, the authors of \cite{rubinstein2004cross,de2005tutorial} introduced the so-called \textit{multi-level cross entropy} method. At each iteration of the algorithm, the cross-entropy problem is solved for a less rare intermediate failure event until reaching the true one. The intermediate events are set from a quantile parameter $\rho$. Several parametric families of densities have been used for this algorithm in the literature: the Gaussian family \cite{rubinstein2004cross,de2005tutorial}, the Gaussian mixture family \cite{kurtz2013cross,geyer2019cross} and the vMFNM family \cite{wang2016cross,papaioannou2019improved} for example. For the following two test cases, we use the classical multi-level cross-entropy but with auxiliary densities parameterised by VAEs (CE-VAE) and in both of them, we set the dimension of the latent space to $d_z = 2$.

At each iteration of the algorithm, we draw $N_{it} = 10^4$ points and the quantile parameter is set to $\rho = 0.25$ for the first example and to $\rho = 0.15$ for the second. We perform $n_{\text{rep}} = 100$ realisations of each estimator to have an estimation of the error. At last, since the Gaussian and the Gaussian mixture families do not perform well at all in high dimension, we will compare the CE-VAE algorithm with the multi-level cross-entropy algorithm with mixtures of vMFNM as auxiliary densities (CE-vMFNM). As said in Section \ref{ssec:is_pres}, the latter algorithm requires the number of components of the mixture, so we will test different setups. For comparison purposes, let us define the following quantities: \begin{itemize}
    \item $N_{\text{tot}}$: the number of calls to the function for each execution of the algorithm,
    \item $\widehat{p}_t^{\text{mean}}$: the mean estimated failure probability over the $n_{\text{rep}}$ realisations,
    \item $\mathrm{COV}\left(\widehat{p}_t\right)$: the coefficient of variation of the estimator over the $n_{\text{rep}}$ realisations,
    \item $\nu_{\text{MC}}$: a coefficient allowing to compare the efficiency of the method compared to the classical Monte Carlo method defined by $\nu_{\text{MC}} = N_{\text{tot}}^{\text{MC}} \left/ N_{\text{tot}}\right.$, where $N_{\text{tot}}^{\text{MC}} = \left(1-p_t\right)\left/\left(p_t\mathrm{COV}\left(\widehat{p}_t\right)\right)\right.$ is the size of the required sample to reach the same coefficient of variation as $\mathrm{COV}\left(\widehat{p}_t\right)$ with the classical Monte Carlo method. If $\nu_{\text{MC}}>1$, the method is more efficient than Monte Carlo and conversely.
    
\end{itemize}

\subsubsection{Four branches}

\begin{figure}
    \centering
    \includegraphics[width=.85\textwidth]{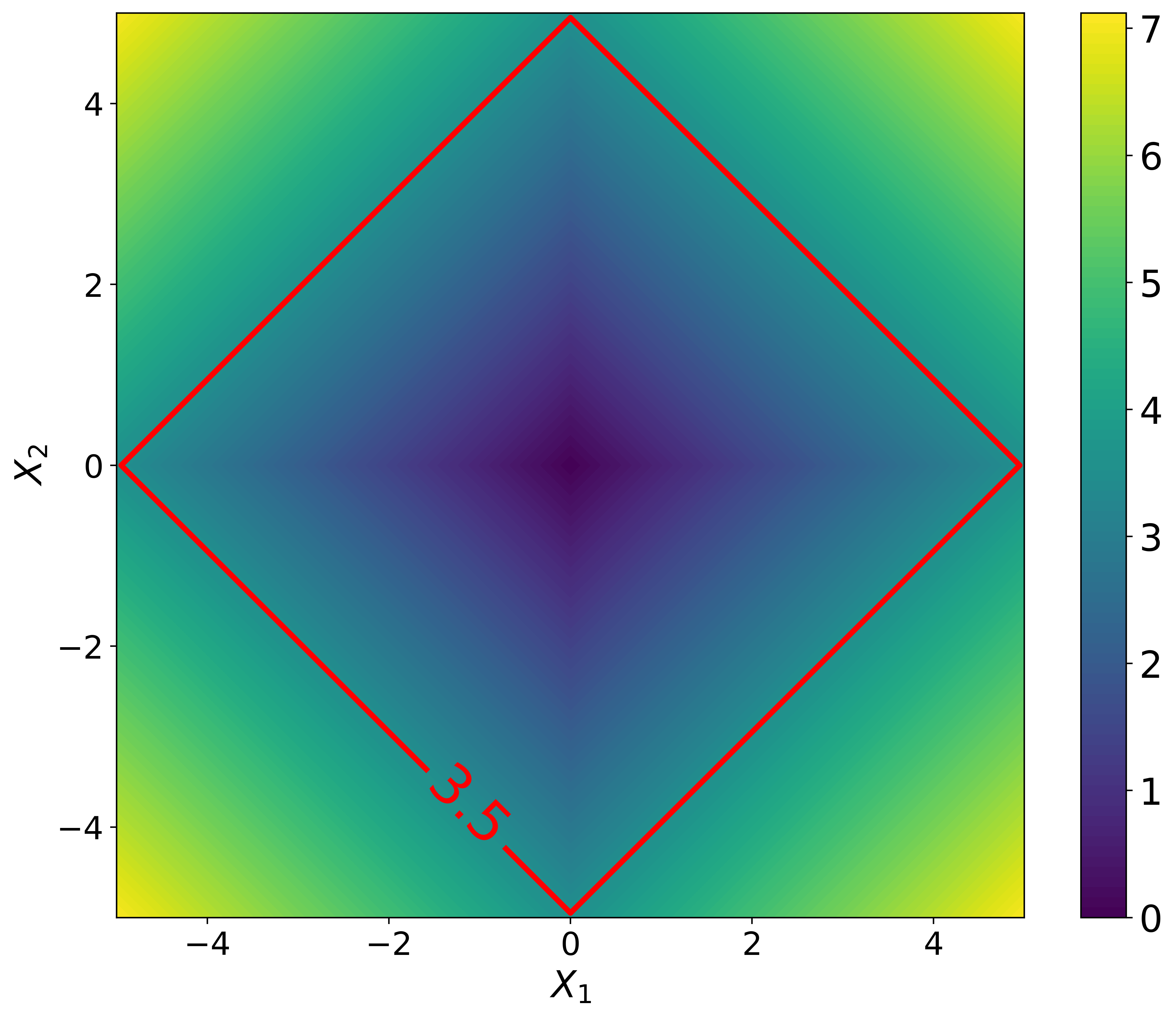}
    \caption{Two-dimensional representation of the function $\psi_1$ as well as the failure limit state corresponding to $t=3.5$ in red.}
    \label{fig:four_branches_2d}
\end{figure}

\begin{figure}
    \centering
    \includegraphics[width=.85\textwidth]{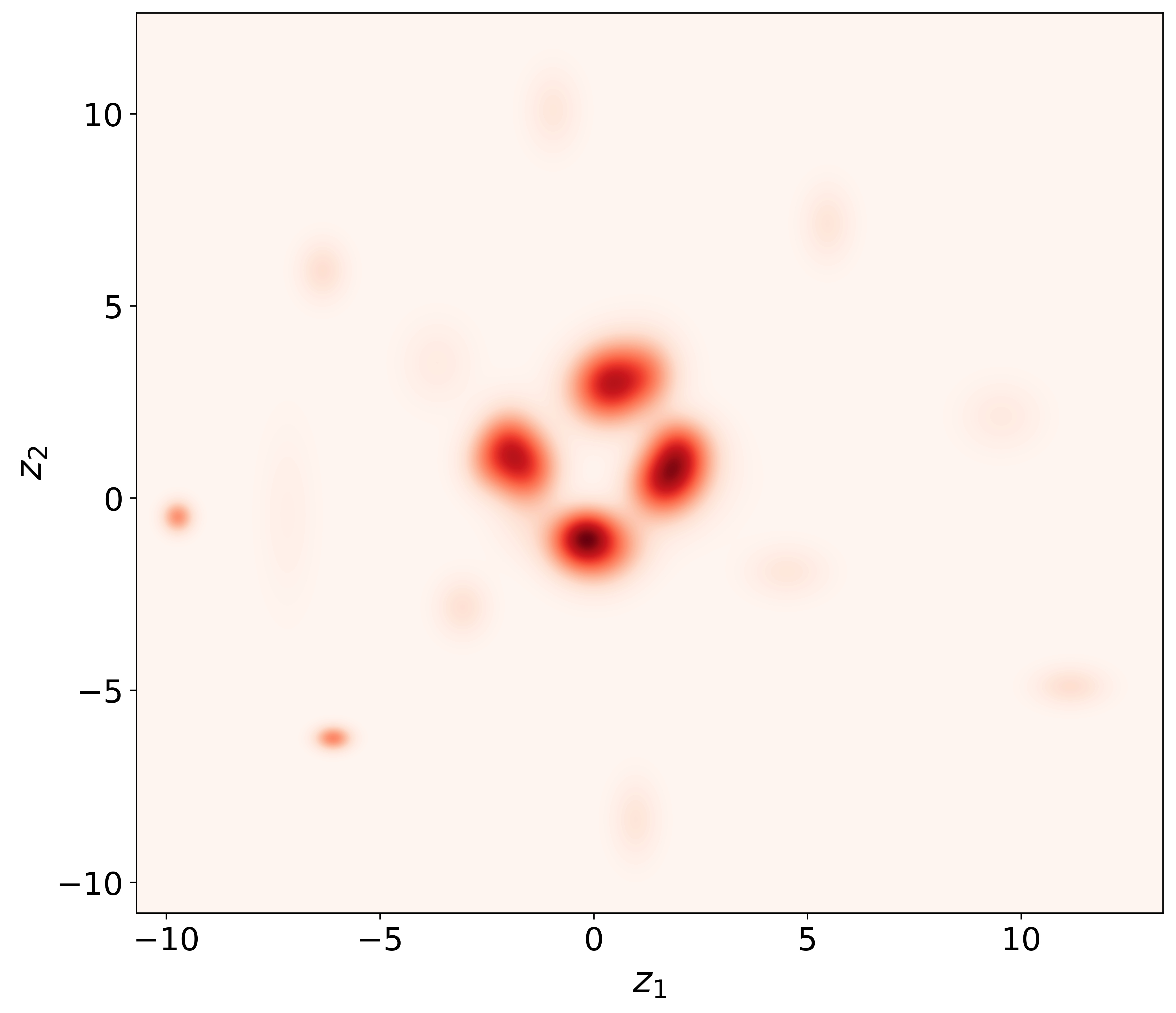}
    \caption{Representation of the two-dimensional PDF of the prior distribution in the latent space associated to the last iteration of one execution of the CE-VAE algorithm.}
    \label{fig:latent_space}
\end{figure}

First, let us consider the analytical problem \cite{chiron2023failure} given for any even dimension $d\geq2$ by: \begin{equation}
    \forall \mathbf{x}\in \mathbb{R}^d, \ \psi_1\left(\mathbf{x}\right) = \min \left\{
    \begin{array}{c}
        \dfrac{1}{\sqrt{d}}\sum_{i=1}^d x_i\\
        -\dfrac{1}{\sqrt{d}}\sum_{i=1}^d x_i\\
        \dfrac{1}{\sqrt{d}}\left(\sum_{i=1}^{d/2} x_i - \sum_{i=d/2+1}^{d} x_i\right)\\
        \dfrac{1}{\sqrt{d}}\left(-\sum_{i=1}^{d/2} x_i + \sum_{i=d/2+1}^{d} x_i\right)
    \end{array}
\right\}.
\end{equation} The input vector associated to this function is the standard Gaussian distribution in dimension $d$. The failure threshold is set to $t=3.5$ such that the theoretical value of the failure probability is $p_t = 9.3\times 10^{-4}$. This problem is challenging because it has 4 failure modes. A graphical representation of this function in dimension $d=2$ as well as the failure limit state are represented in Figure \ref{fig:four_branches_2d}.Here, we set the dimension of the problem to $d=100$, and the numerical performances of the algorithms are given in Table \ref{table:four_branches_numres}. By comparing both the coefficient of variation and the coefficient $\nu_{\text{MC}}$ associated to each estimator, we can see that the proposed CE-VAE algorithm provides better performances than the CE-vMFNM algorithm here, and so without any prior knowledge on the number of failure modes. Figure~\ref{fig:latent_space} represents PDF of the prior distribution in the latent space at the last iteration for one execution of the CE-VAE algorithm. One can clearly see that the four failure modes are well represented in it, therefore when generating new points the four modes will be represented. Moreover, one can note that the choice of the number of components in the mixture for the CE-vMFNM algorithm is crucial. Indeed, since the current problem has 4 failure modes, we need at least 4 components in the mixture in order to get good results. The algorithm for 3 components did not even converge because for each execution, it reached the maximal number of iterations. At last, unreported visual inspection shows that the empirical distribution of the $n_{\text{rep}} = 100$ realisations of the CE-vMFNM4 algorithm has a heavy tail, and therefore explains the observed error.

\begin{table}[h!]
\caption{Comparison of the CE-VAE algorithm with the CE-vMNFM algorithm with different numbers of components for the four branches problem. The number after the ``CE-vMFNM'' acronym represents the number of components of the mixture given as an algorithm input.}
\centering
\begin{tabular}{|c||c|c|c|c|} 
 \hline
& CE-VAE & CE-vMFNM3 & CE-vMFNM4 & CE-vMFNM5\\
 \hline\hline
 $N_{\text{tot}}$ & 40000 & 200000 & 50000 & 50000\\
 \hline
$\widehat{p}_t^{\text{mean}}$ & $9.310\times10^{-4}$ & $1.971\times10^{-3}$ & $9.835\times10^{-4}$ & $9.315\times10^{-4}$ \\ 
\hline
$\mathrm{COV}\left(\widehat{p}_t\right)$ & $5.31\%$ & $989.5\%$ & $31.3\%$& $7.56\%$ \\
\hline
$\nu_{\text{MC}}$ & $9.54$ & $5.49\times10^{-5}$ & $2.19\times10^{-1}$ & $3.76$  \\
\hline
\end{tabular}
\label{table:four_branches_numres}
\end{table}  

\subsubsection{Duffing oscillator}

The second example is the Duffing oscillator introduced in \cite{zuev2009advanced}, and we consider here its writing discretised in the frequency domain as in \cite{shinozuka1991simulation,papaioannou2019improved}. The function of interest is the maximal displacement of the oscillator at $a_{\text{max}} = 2\mathrm{sec}$ given by: \begin{equation}
    \psi_2\left(\mathbf{x}\right) = \min\left\{u_1 - u\left(a_{\text{max}}\right),u\left(a_{\text{max}}\right)-u_2\right\},
\end{equation} where $u_1 = 0.1$ and $u_2 = -0.06$, and where the displacement $u\left(a\right)$ of the oscillator satisfies for all $a\geq 0$:\begin{equation}
    m \ddot u\left(a\right) + c \dot u\left(a\right) + k\left(u\left(a\right) + \gamma u\left(a\right)^3\right) = -m\sigma\sum_{i=1}^{d/2} \left(X_i\cos\left(\omega_i a\right) + X_{d/2+i}\sin\left(\omega_i a\right)\right), 
\end{equation} with $m=1000 \ \mathrm{kg}$, $c = 200\pi \ \mathrm{Ns/m}$, $k=1000\left(2\pi\right)^2 \ \mathrm{N/m}$, $\gamma = 1 \ \mathrm{m}^{-2}$, $w_i = i\Delta\omega$, $\Delta\omega = 30\pi/d$ and $\sigma = \sqrt{0.01\Delta\omega}$. Moreover, the initial conditions are set to $u\left(0\right) = 0$ and $\dot u\left(0\right) = 1.5$, the random variable $\mathbf{X}$ follows a standard Gaussian distribution of dimension $d=200$ and the failure threshold is set to $t=0$. It is possible to verify that there are then two failure modes. The reference value of the failure probability is $p_{\text{ref}} = 4.28\times 10^{-4}$ and it has been computed with a large Monte Carlo estimation of size $10^6$ \cite{chiron2023failure}.

\begin{table}[h!]
\caption{Comparison of the CE-VAE algorithm with the CE-vMNFM algorithm with different number of components on the Duffing oscillator problem. The number after the ``CE-vMFNM'' acronym represents the number of components of the mixture given as an algorithm input.}
\centering
\begin{tabular}{|c||c|c|c|c|} 
 \hline
& CE-VAE & CE-vMFNM1 & CE-vMFNM2 & CE-vMFNM3\\
 \hline\hline
 $N_{\text{tot}}$ & 30000 & 200000 & 40000 & 40000 \\
 \hline
$\widehat{p}_t^{\text{mean}}$ & $4.27\times10^{-4}$ & $1.27\times10^{-1}$ & $4.25\times10^{-4}$ & $4.26\times10^{-4}$ \\ 
\hline
$\mathrm{COV}\left(\widehat{p}_t\right)$ & $8.69\%$ & $2317.4\%$ & $3.72\%$& $2.51\%$ \\
\hline
$\nu_{\text{MC}}$ & $10.3$ & $2.17\times10^{-5}$ & $42.2$ & $92.8$ \\
\hline
\end{tabular}
\label{table:oscillator_numres}
\end{table} 

The numerical performances of the algorithms are given in Table \ref{table:oscillator_numres}. By comparing once more both the coefficient of variation and the coefficient $\nu_{\text{MC}}$ associated to each estimator, the CE-VAE algorithm provides satisfying performances, but the CE-vMFNM2 and CE-vMFNM3 algorithms outperform it. Indeed, their coefficient of variation is between two and three times smaller than the one of the CE-VAE algorithm and although the CE-vMFNM algorithms require one more iteration to converge, their associated coefficient $\nu_{\text{MC}}$ is much better. The current problem is probably very well suited for this family of auxiliary distributions and can explain these very good results in such high dimension. However, as already said in the previous example, contrary to the proposed CE-VAE algorithm, the CE-vMFNM algorithm requires the knowledge of the number of components in the mixture, which can often be a limitation. Here, at least two components are required because we can observe that the CE-vMFNM1 algorithm does not work at all since it reaches the maximal number of iteration for each execution. The proposed training procedure of the VAE allows to identify both failure modes without any prior knowledge and to get satisfying results anyway.

\section{Conclusion}
\label{sec:conclusion}

In the present article, we are interested in probability density estimation with weighted samples, i.e. the estimation of a target distribution $g^*$ by using a sample drawn from another distribution $f$. It is a major topic of interest in statistics which has many applications, some of them presented in Section \ref{sec:numerical_results}. We suggest to approach $g^*$ by a distribution parameterised by a VAE. To do so, we extend the existing VAE framework to the case of weighted samples, by introducing the new objective function $\text{wELBO}$ to maximise in order to learn the best parameters of the neural networks. Even if the corresponding distribution theoretically belongs to a parametric family, its characteristics make it closer to a non-parametric model. Despite the very high number of parameters to estimate, this family is much more efficient in high dimension than the classical Gaussian or Gaussian mixture families. Moreover, in order to add flexibility to the model, and more precisely to be able to learn multimodal distributions, we use a learnable prior distribution for the latent variable called VampPrior. We also introduce a new pre-training procedure for the VAE in order to find good starting points $\left(\boldsymbol{\phi}^{(0)},\boldsymbol{\theta}^{(0)},\boldsymbol{\lambda}^{(0)}\right)$ for the maximisation of $\text{wELBO}$ and to prevent as much as possible the posterior collapse phenomenon to happen. At last, we illustrate and discuss the practical interest of the proposed method. We first describe the classical IS framework for estimating an expectation and we show how to exploit the resulting distribution to that purpose. Then, we apply the proposed procedure in an adaptive IS algorithm for drawing points according to a target distribution. Both examples in dimension 10 and 20 show that a VAE has a larger flexibility and is more likely to catch the specificity of a distribution than a Gaussian or a Gaussian mixture distribution. Finally, we introduce the proposed method in an adaptive IS algorithm for estimating a failure probability in high dimension. We observe that the resulting estimation is always quite accurate without any prior knowledge on the form of the problem, in opposition to the vMNFM family.

It is important to note that the computational time of the training of neural networks, and so of a VAE, is growing with the number of training points and the input dimension. As a comparison, the training time of the other parametric families used in this article is almost instantaneous whereas one execution of the proposed procedure can last from less than one minute to 5 minutes on a CPU. However, once the VAE is trained, the generation of new points is instantaneous, and when considering time-expensive black-box functions, the training time of a VAE becomes negligible. Moreover, the suggested architecture in Figure \ref{fig:vae_archi} is not the only possibility. Indeed, we suggest here a unique architecture quite simple to implement but general enough to be adapted for every considered problem. However, if it is necessary, an interested user can modify the suggested architecture to be more adapted to its data. 

Finally, this article shows that a VAE is a powerful tool for density estimation with or without weighted samples, and we apply it for adaptive IS. Then, one can investigate the use of VAEs for others algorithms or uncertainty quantification problems requiring high-dimensional density estimation, such as an improved adaptive IS algorithm for failure probability estimation \cite{papaioannou2019improved} or some MCMC methods \cite{chib1995understanding,au2001estimation} for example.

\section*{Acknowledgements}

The first author is enrolled in a Ph.D. program co-funded by \textit{ONERA – The French Aerospace Lab} and \textit{Toulouse III - Paul Sabatier University}. Their financial supports are gratefully acknowledged.

\bibliography{main}
\bibliographystyle{tmlr}

\end{document}